\documentclass[sigconf,nonacm]{acmart}

\AtBeginDocument{%
  \providecommand\BibTeX{{%
    \normalfont B\kern-0.5em{\scshape i\kern-0.23em b}\kern-0.8em\TeX}}}

\usepackage{marvosym}
\usepackage{float}
\usepackage{graphicx}
\usepackage{placeins}
\begin{document}

\title{What Makes Good Few-shot Examples for Vision-Language Models?}

\author{Zhaojun Guo}
\authornote{Both Zhaojun Guo and Jinghui Lu contributed equally to this research.}
\email{22110240087@m.fudan.edu.cn}
\orcid{0000-0002-6833-9380}
\affiliation{%
  \institution{Fudan University}
  \city{Shanghai}
  \country{China}
}

\author{Jinghui Lu}
\email{lujinghui@bytedance.com}
\authornotemark[1]
\orcid{0000-0001-7149-6961}
\affiliation{%
  \institution{Bytedance}
  \city{Shanghai}
  \country{China}
}

\author{Xuejing Liu}
\email{liuxuejing@sensetime.com}
\affiliation{%
  \institution{Sensetime Technology}
  \city{Shanghai}
  \country{China}
}

\author{Rui Zhao}
\email{zhaorui@sensetime.com}
\affiliation{%
  \institution{Sensetime Technology}
  \city{Shanghai}
  \country{China}
}

\author{Zhenxing Qian\textsuperscript{\Letter}}
\email{zxqian@fudan.edu.cn}
\orcid{}
\affiliation{%
  \institution{Fudan University}
  \city{Shanghai}
  \country{China}
}

\author{Fei Tan\textsuperscript{\Letter}}
\email{tanfei@sensetime.com}
\affiliation{%
  \institution{Sensetime Technology}
  \city{Shanghai}
  \country{China}
}


\begin{abstract}
Despite the notable advancements achieved by leveraging pre-trained vision-language (VL) models through few-shot tuning for downstream tasks, our detailed empirical study highlights a significant dependence of few-shot learning outcomes on the careful selection of training examples—a facet that has been previously overlooked in research. In this study, we delve into devising more effective strategies for the meticulous selection of few-shot training examples, as opposed to relying on random sampling, to enhance the potential of existing few-shot prompt learning methodologies. To achieve this, we assess the effectiveness of various Active Learning (AL) techniques for instance selection, such as Entropy and Margin of Confidence, within the context of few-shot training. Furthermore, we introduce two innovative selection methods — {\textbf{Repre}sentativeness (REPRE)} and Gaussian \textbf{Monte Carlo} (Montecarlo) — designed to proactively pinpoint informative examples for labeling in relation to pre-trained VL models. Our findings demonstrate that both REPRE and Montecarlo significantly surpass both random selection and AL-based strategies in few-shot training scenarios. The research also underscores that these instance selection methods are model-agnostic, offering a versatile enhancement to a wide array of few-shot training methodologies.

\end{abstract}





\maketitle

\section{Introduction}

\begin{table}[ht]
\footnotesize
\centering
\setlength{\tabcolsep}{2.5mm}
\caption{An examination of the CoOp's 2-shots prompt learning performance using different few-shot training sets.}
\label{tab:prestudy}
{%
\begin{tabular}{lllll}
\toprule
\textbf{Dataset}       & \multicolumn{3}{l}{\textbf{Accuracy}} & \textbf{Std}   \\ \midrule
              & Seed 1   & Seed 2  & Seed 3  &       \\ \cline{2-4} \\ [-4pt]
EuroSAT       & 66.70     & 56.90    & 59.90    & 5.02  \\
FGVCAircraft & 27.60     & 3.60     & 26.90    & 13.65 \\ \bottomrule
\end{tabular}%
}
\end{table}
With the development of the deep learning, significant enhancements in vision-related tasks, including classification~\cite{Lanchantin_2021_CVPR}, detection~\cite{Xie_2021_ICCV}, temporal action detection~\citet{feng2023refinetad} and segmen-tation\cite{Li_2022_CVPR, li2023catr}, have been achieved through the introduction of superior neural architectures. These advancements have led to notable improvements in task accuracy, demonstrating the impact of innovative design in neural networks and framework development on the field of computer vision.
Building on these seminal advances in computer vision, vision-language models like CLIP and Maple have bridged the gap between visual perception and language processing, marking a new era of AI capabilities. 

Vision-language models~\citeN{CLIP, liu2023enhancing} have recently demonstrated str-ong generalization performance on downstream tasks. Basically, these models are pretrained on web-scale datasets comprising image-text pairs, enabling them to understand and interpret complex visual and linguistic cues in unison. Then, various few-shot or zero-shot techniques are applied to adapt the model to downstream tasks. For example,~\citet{CLIP} designed 80 text prompt templates to fully leverage CLIP's zero-shot capabilities, demonstrating that representations learned from the pre-trained vision-language model (e.g.,~\cite{CLIP}) can be transferred to a wide range of downstream tasks. 
Following~\citet{CLIP}, a branch of work~\citeN{Maple, CoOp, gondal2024domain, peng2023sgva} investigate optimal continuous prompt templates that are trained with a handful of labelled examples to integrate downstream knowledge into pre-trained CLIP rather than using hand-crafted templates, achieving impressive few-shot performance.

\citet{CoOp} introduced a novel approach for tailoring CLIP-like vision-language models to downstream image recognition tasks through the use of customized prompts. Building upon this groundwork,~\citet{zhou2022conditional} advanced the concept further by developing a lightweight neural network dedicated to an input-conditional token (vector) generation for each image, enhancing the model's adaptability and context sensitivity. Additionally,~\citet{Maple} unveiled a method aimed at refining the synergy between vision and language modalities. This technique focuses on improving the alignment between visual and linguistic representations, thereby enhancing the model's overall performance in processing and understanding multimodal content. 

However, previous research has primarily focused on investigating methods for optimizing prompts, neglecting the significance of the few-shot training examples in enhancing prediction performance.\\ 
In practice, the performance of those few-shot prompt learning approaches tends to vary through different choices of few-shot examples. As exemplified in Table~\ref{tab:prestudy} when using CoOp~\cite{CoOp} as a few-shot learner, the results exhibit high sensitivity to the specific examples utilized.
Mainly, in \textit{FGVCAircraft}, seed 1 achieves 27.60 accuracy while only 3.60 when using seed 2.

Our research is dedicated to meticulously analyzing the variability in performance observed across few-shot prompt learning approaches (i.e., few-shot learners), attributable to differentiating selections of few-shot examples. This variability raises the question of \textit{whether a more systematic data selection strategy could outperform the random sampling commonly employed in existing studies. }


In pursuit of an answer, we explore the efficiency of two well-known Active Learning strategies—Entropy and Margin of Confidence—within the few-shot learning framework, noting that active learning is renowned for its strategic selection of data~\cite{settles2009active,ren2021survey}. The Entropy and Margin of Confidence strategies focus on identifying examples about which the pre-trained model demonstrates the greatest uncertainty. This approach is predicated on the hypothesis that training on these uncertain examples could significantly enhance the model's learning efficiency in few-shot scenarios. The intricacies of these methods and their implementation will be thoroughly explored in Section~\ref{subsec:active_learning}. However, our findings indicate these methods offer minimal improvement or even negative impact as compared to random selection (Section~\ref{subsec: overall_performance}). Therefore, we introduce two innovative data selection approaches specifically designed for vision-language (VL) models --- \textbf{Repre}sentativeness (REPRE) and Gaussian \textbf{Monte Carlo} (Montecarlo). 

The Gaussian Monte Carlo method, by assuming that the visual encoder becomes more robust with images it frequently encounters during pre-training, we select less familiar examples to quickly bridge the knowledge gap. The ``familairity'' of an example with respect to a pre-trained VL model is gauged through the lens of its distance from modified versions with added Gaussian noise, this method seeks to pinpoint the examples that offer the most valuable insights for the pre-trained model. This innovative approach employs algorithms that use repeated random sampling from a Gaussian distribution to approximate solutions for complex problems, such as a novel metric for example selection. This method is similar to the Monte Carlo process and is therefore named Gaussian Monte Carlo. A detailed explanation of its rationale and implementation is provided in Section~\ref{subsec:Montecarlo}.

On the other hand, the Representativeness strategy selects examples that are deemed most emblematic of the broader dataset, aiming to ensure that the few-shot examples provide a comprehensive overview of the data's diversity, which will be elaborated in Section~\ref{subsec:repre}. 
In summary, the main contributions of this work are as follows:
\begin{itemize}
    \item To the best of our knowledge, we are the first to explore the effectiveness of the example selection methods for the few-shot training of pre-trained vision-language models.
    \item We explore the efficiency of two active learning AL-based data selection strategies in a few-shot context, showing that simply incorporating AL methods into few-shot scenarios proves to be ineffective.
    \item We propose two novel instance selection methods: Gaussian \textbf{Monte Carlo} (Montecarlo) and \textbf{Repre}sentativeness (REPRE). Extensive experimental results demonstrate that these proposed methods can enhance the performance of state-of-the-art few-shot prompt learning methods. This improvement is significant when compared to traditional strategies, i.e.,random selection, leading to superior prediction performance.
\end{itemize}

\section{Related work}
\subsection{Vision-Language Models}
Vision-Language (VL) pre-trained models, like CLIP~\cite{CLIP}, have emerged as powerful tools for acquiring versatile visual representations, which can be effectively transferred to a wide array of downstream classification tasks through the use of carefully crafted prompts. Building upon the foundation laid by CLIP, there has been a surge of research efforts~\cite{CoOp}, with more references to be added, aimed at enhancing the adaptability and effectiveness of these models for specific downstream datasets. This is achieved by adopting continuous prompts rather than relying on fixed manual templates. Among these innovative methods, CoOp~\cite{CoOp} stands out as a pioneering approach that advocates for the learning of dynamic language prompts, especially in few-shot learning environments, quickly becoming a cornerstone technique in the field. Adding to this landscape,~\citet{Maple} introduced MaPLe, an advanced methodology that augments language prompts with additional visual prompts, establishing a comprehensive multi-modal prompt learning framework that achieves unparalleled few-shot learning performance. ~\citet{liu2023deeply} introduce a method called Deeply Coupled Cross-modal Prompt Learning (DCP), which is based on the CLIP architecture. DCP effectively integrates the interaction between vision and language using a Cross-Modal Prompt Attention (CMPA) mechanism. This approach facilitates a dynamic exchange of respective representations through a robustly interconnected multi-head attention module, enhancing the system's capabilities progressively and substantially. ~\citet{wu2023feature} present an adapter designed to boost the efficiency of CLIP adaptation in few-shot learning. It works by accentuating the distinctions between the fine-tuned CLIP features and the original CLIP features. In our research, we apply our novel data selection strategies in conjunction with CoOp and MaPLe, leveraging their widespread recognition and proven effectiveness to further push the boundaries of what's possible in few-shot prompt learning.

\subsection{Instance Selection}
To our knowledge, our study marks the inaugural exploration into how data selection affects few-shot training within the realm of pre-trained vision-language models. While the concept of selecting optimal few-shot training examples has been scrutinized within the fields of natural language processing (NLP) and computer vision (CV) independently, it has not been jointly examined in the context of multi-modal learning.
\paragraph{Natural Language Processing}
In Natural Language Processing, the pioneering work by~\citet{NLP_selection} highlighted the critical role of instance selection in enhancing few-shot text generation. Their approach, which involves clustering the training set via K-means and selecting the centroid of each cluster as a representative sample, deviates from the traditional random selection methodology. Furthermore,~\citet{liu-etal-2022-makes} introduced a strategy that prefers examples closest to the test instances in the embedding space for few-shot learning. However, their focus was on in-context learning for decoder-only architectures, such as those in the GPT series, diverging from the continuous prompt learning approach.

\paragraph{Computer Vision}
Concurrently, in the CV domain, the use of active learning techniques for selectively populating few-shot support sets has been explored by~\citet{Active}, while~\citet{zhou2020learning} have devised a method for choosing optimal base classes for few-shot learning. These initiatives, though innovative, are primarily tailored to CV models within a meta-learning framework and do not extend to multi-modality models guided by linguistic context. Additionally,~\citet{xu2022generating} explored the generation of supplementary training examples to bolster few-shot learning performance, a strategy that falls outside the purview of our investigation. Despite these methods showing promise in single-modality settings, they have yet to be applied or tested within multi-modal contexts. Given the pronounced disparities between multi-modal and mono-modal learning environments, including differences in model architectures, inference techniques, and evaluation protocols, there's a compelling need for further research to unravel the nuances of few-shot example selection in pre-trained vision-language models.

\section{Methods}
In this section, we outline the formulation of the research problem. Following this, we detail two active learning data selection methods and our approach to adapting them for few-shot contexts. We then provide an in-depth elaboration of the two proposed data selection methods: Gaussian Monte Carlo and Representativeness.

\subsection{Problem Formulation}
Following the few-shot learning scenario, we denote the unlabeled data as $u_1,u_2,...,u_n$, where $n$ is the data size. In the visual recognition tasks, the instances can be images, and we annotate them and then fine-tune our pre-trained vision-language models based on the annotated data. The goal is to \textit{find the most valuable labeled data that can improve prediction performance to the largest extent given a certain label budget}.

\subsection{Entropy and Margin of Confidence} \label{subsec:active_learning}
We first investigate two classic instance selection methods from AL, espeically uncertainty-based method, that is, Entropy selection strategy~\cite{Active} and Margin of Confidence~\cite{settles2009active,ren2021survey}. In these two methods, examples that models are most uncertain about are considered the most informative ones and thus to be labeled. Entropy characterizes uncertainty as the entropy of the model's prediction distribution, which is computed as follows:

\begin{equation}
    E = \sum_{i=1}^N P_i \ln P_i
\end{equation}\label{eq:entropy}

\noindent where $E$ is the entropy, $P_i$ is the model prediction probability of the $i^{th}$ class. Examples with the highest entropy are selected to be labeled. While Margin of Confidence interprets uncertainty as the difference between the two most confident predictions, which is computed by: 

\begin{equation}
    E = P_{max} - P_{max-1}
\end{equation}\label{eq:margin}

\noindent where $P_{max}$ is the most confident prediction probability and $P_{max-1}$ is the second most confident prediction probability). In Margin of Confidence, examples with minimal difference are regarded as the most uncertain cases and are selected to be labeled. In this work, we normalize CLIP zero-shot scores to estimate the prediction probability for each class. 

While showing promise as data selection strategies in active learning scenarios, our experimental findings reveal that these two methods fall short in few-shot contexts with VL models (Section~\ref{subsec: overall_performance}). Therefore, in subsequent sections, we will introduce the premise and the workflow of the two proposed methods: Gaussian Monte Carlo and Representativeness.

\subsection{Gaussian Monte Carlo (Montecarlo)}\label{subsec:Montecarlo}
\begin{figure*}[ht]
    \centering 
    \includegraphics[width=0.9\textwidth]{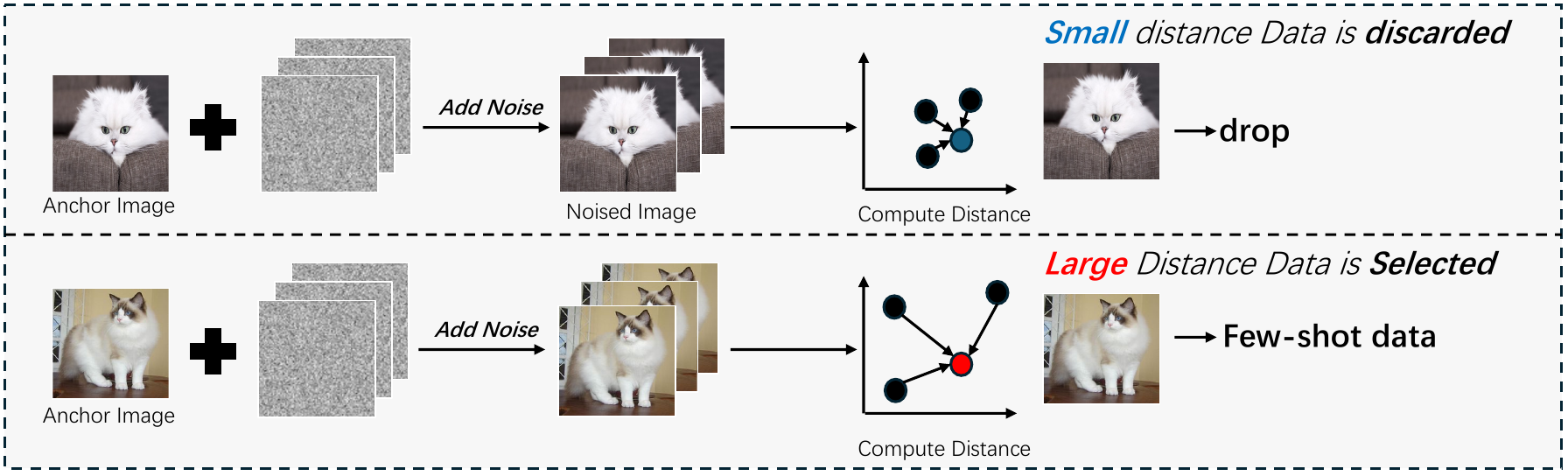}   
     \caption{Illustration of computation of Montecarlo score, the image with a larger score is selected.} 
       \Description{Illustration of computation of Montecarlo score}
    \label{Fig.main1} 
\end{figure*}


We propose a \textbf{Gaussian Monte Carlo} instance selection method which considers the examples that the pre-trained VL model is most unfamiliar with as the most useful examples to be labeled. The familiarity here refers to the extent to which the model encountered similar images during pre-training. 

In our study, we suggest that the magnitude of familiarity a model has with an image can be assessed by the average distance between the representations of the original image and its Gaussian noised counterparts. This method is based on the assumption that: \textit{the pre-trained VL model demonstrates greater robustness with images it recognizes more familiarly. In other words, if a VL model frequently encounters an image or similar ones during pre-training, then, even when the image is perturbed, the modified image should remain close to the anchor image in the embedding space.} Consequently, the Montecarlo pipeline is shown as follows:

Firstly, we will generate gaussiaon noised image of each original image (i.e., anchor image). Gaussian noise is statistical noise with a probability density function of the zero means normal distribution. We can obtain the output pixel for each input pixel by adding it to a random number that matches the Gaussian distribution. The processing is formulated as:

\begin{equation}
    I'= \frac{1}{M \times N} \sum_{i=0}^{M} \sum_{j=0}^{N}\left(I_{i j}+Z\right) 
\end{equation}

\begin{equation}
    Z=f(x)=\frac{1}{\sigma\sqrt{2\pi}}\exp (-\frac{(x-\mu)^2}{2\sigma^2})
\end{equation}

\noindent where $I$ is the original image and $i$ and $j$ are the coordinates of the pixel to which the noise is applied. We randomly generate a Gaussian number $Z$, which is added to each pixel. $I'$ is the Gaussian noised image. In this process, we will generate tens of gaussian noised images for each anchor image.

Then we calculate the mean distance $D(I, I')$ between the anchor image (i.e., original image) and Gaussian noised image. The objective of our proposed method is to identify the most informative instances, thereby enabling the pre-trained model to acquire the most valuable samples, as illustrated in Figure~\ref{Fig.main1}. We achieve this by selecting the top $K$ anchor images that exhibit the greatest distance between the noisy and original images in the embedding space. As described previously, anchor images that their noised images are closer are recognized as more familiar by pre-trained VL models. Conversely, those positioned further away are considered less familiar, making them more valuable for selection. Here, $K$ represents the number of shots.

\subsection{Representativeness}\label{subsec:repre}
In this method, the goal is to identify the most representative samples for use in few-shot learning.  In our study, ``representativeness'' refers to the degree to which a data example closely resembles or exemplifies the central tendencies or characteristics of its respective class or category within an embedding space. This concept hinges on the idea that \textit{each class or category of data can be represented by a centroid in an embedding space. A data example's representativeness is quantitatively measured by its distance to its class centroid in this space; the closer a data example is to the centroid, the more representative it is considered.} This measure allows for the selection of the most characteristic examples of each class, which is crucial for few-shot learning scenarios where the goal is to achieve high model performance with a limited number of training samples.

Thus, we consider the representativeness is determined by measuring the distance from a candidate data example to the class centroid within the embedding space. We first compute the mean embedding for each class by validation set. Then we calculate the distance between each example and the centroid as the score indicating the representativeness of this example. The closer an image is to the centroid of its class, the more representative it is considered. Consequently, the top $K$ most representative images will be selected.

\begin{figure*}[htbp]
\centering
\begin{tabular}{ccc}
\includegraphics[width=0.23\textwidth]{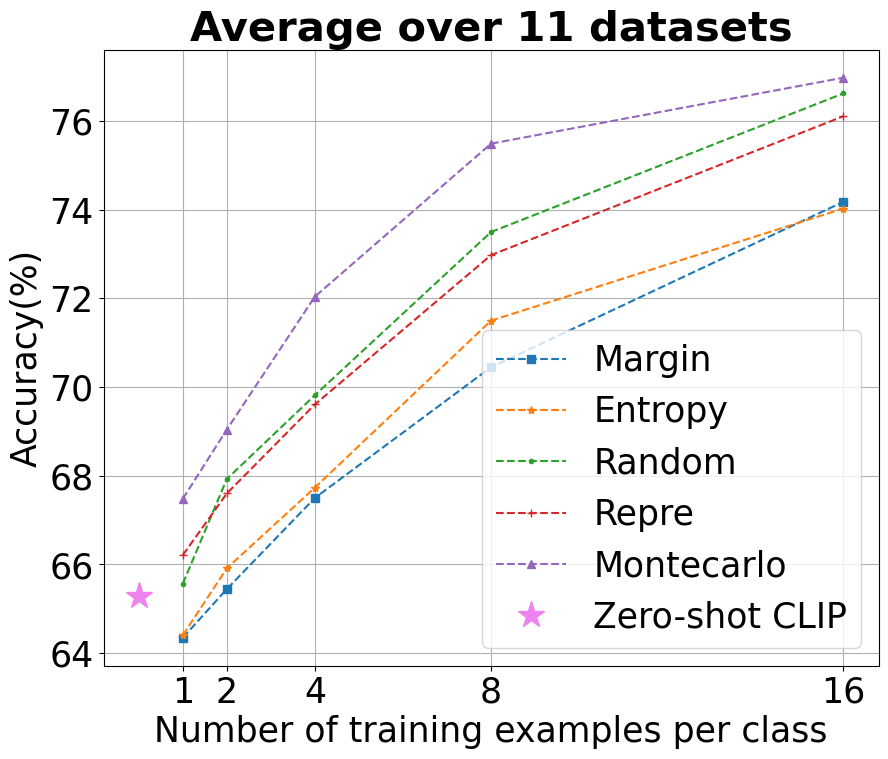} &
\includegraphics[width=0.23\textwidth]{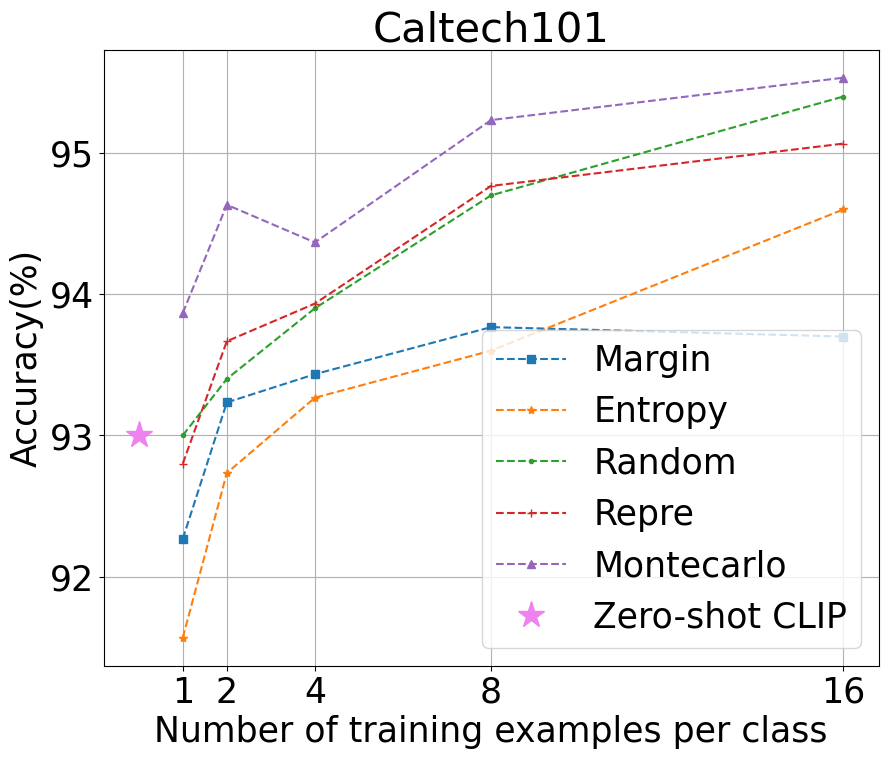} &
\includegraphics[width=0.23\textwidth]{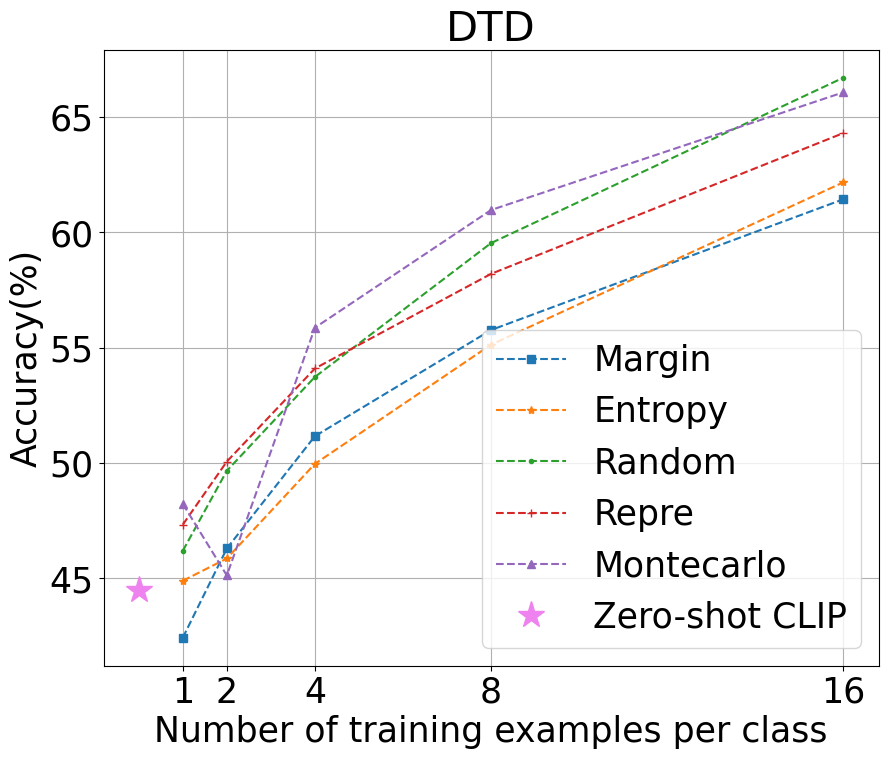} \\
\includegraphics[width=0.23\textwidth]{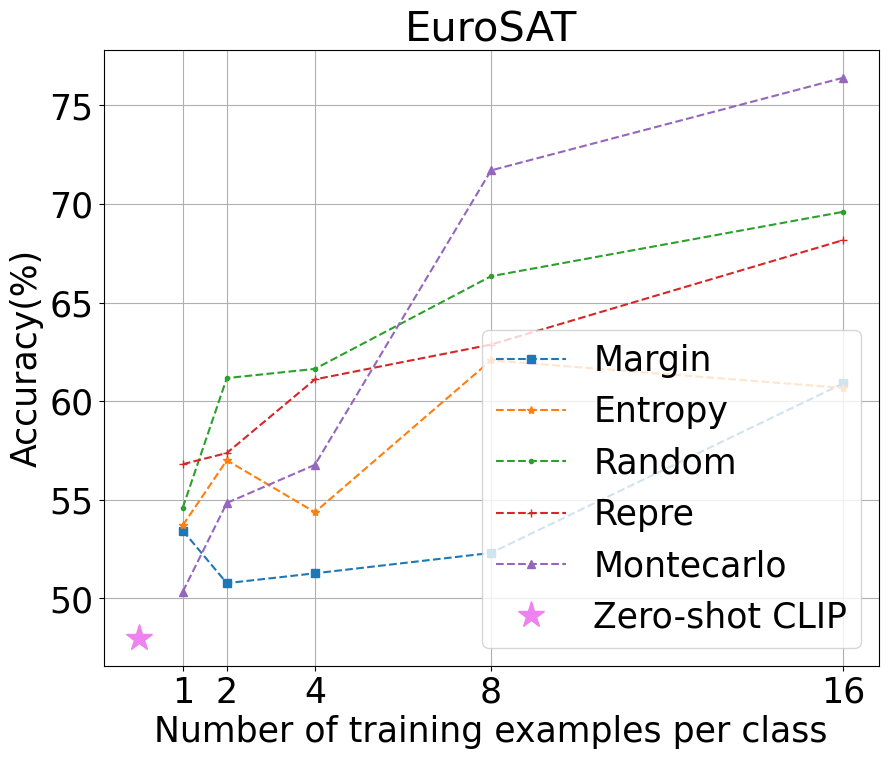} &
\includegraphics[width=0.23\textwidth]{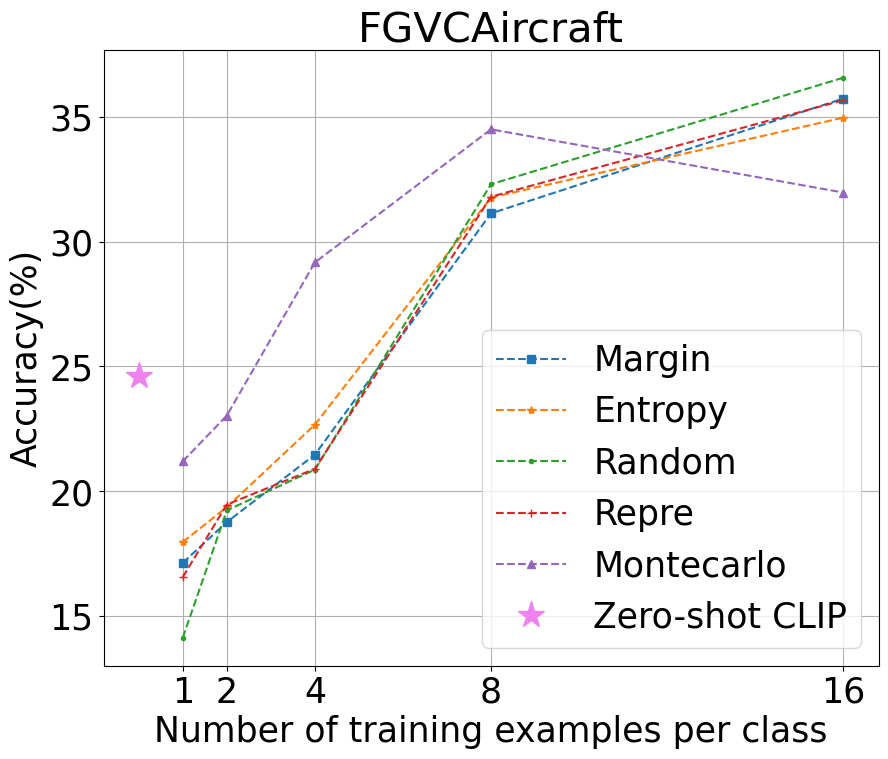} &
\includegraphics[width=0.23\textwidth]{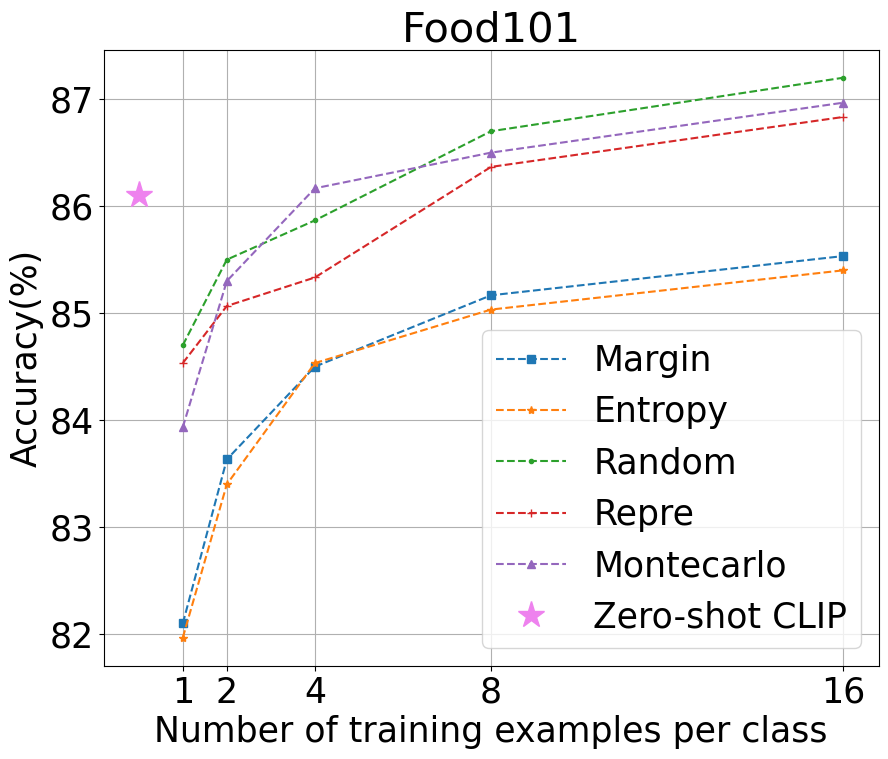} \\
\includegraphics[width=0.23\textwidth]{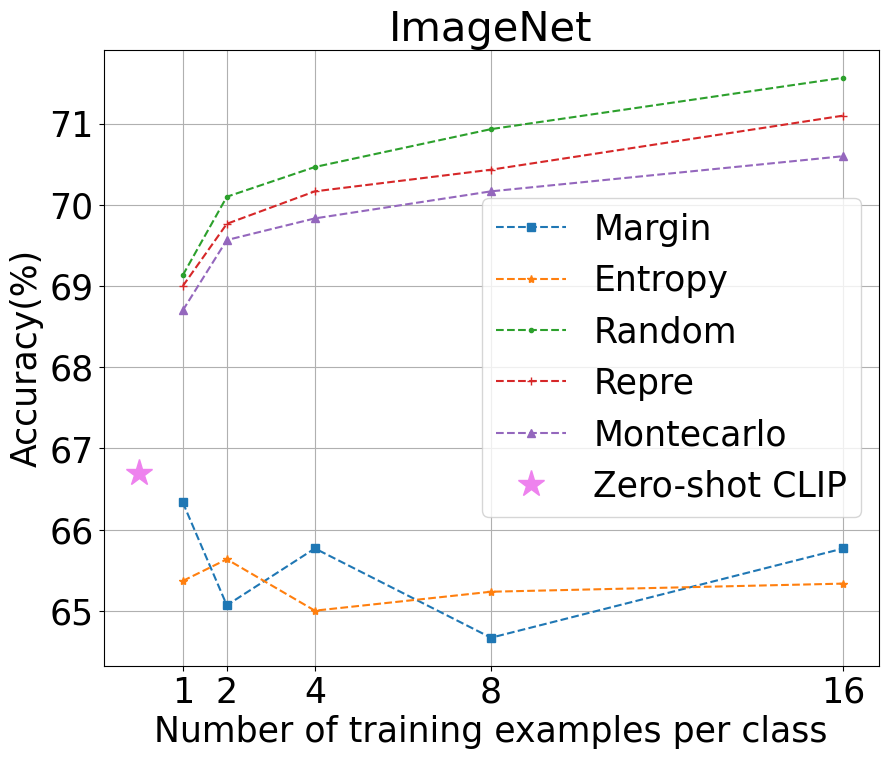} &
\includegraphics[width=0.23\textwidth]{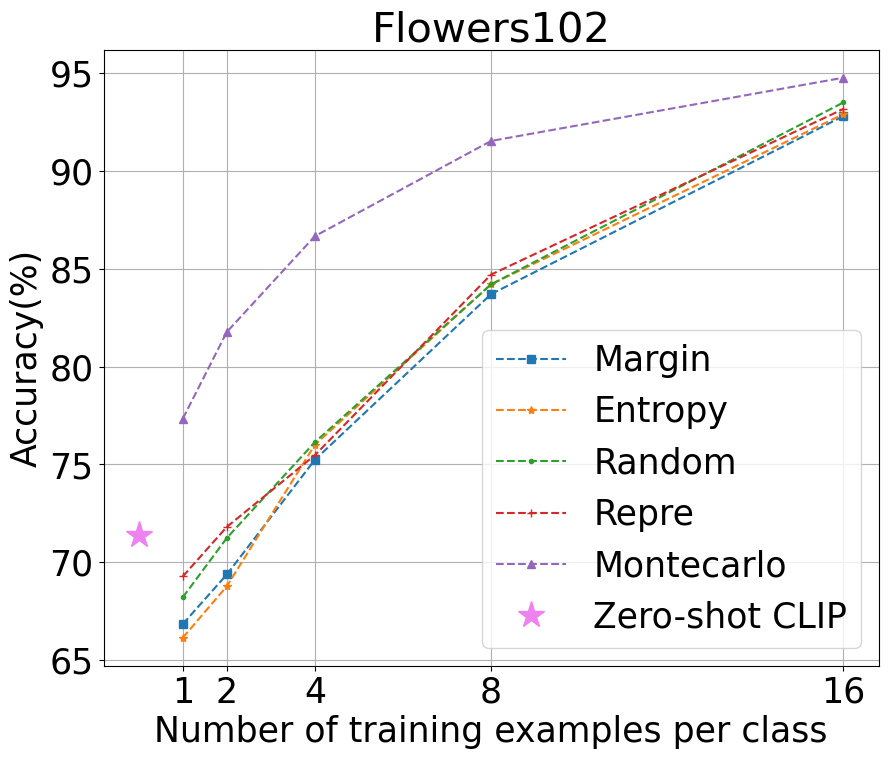} &
\includegraphics[width=0.23\textwidth]{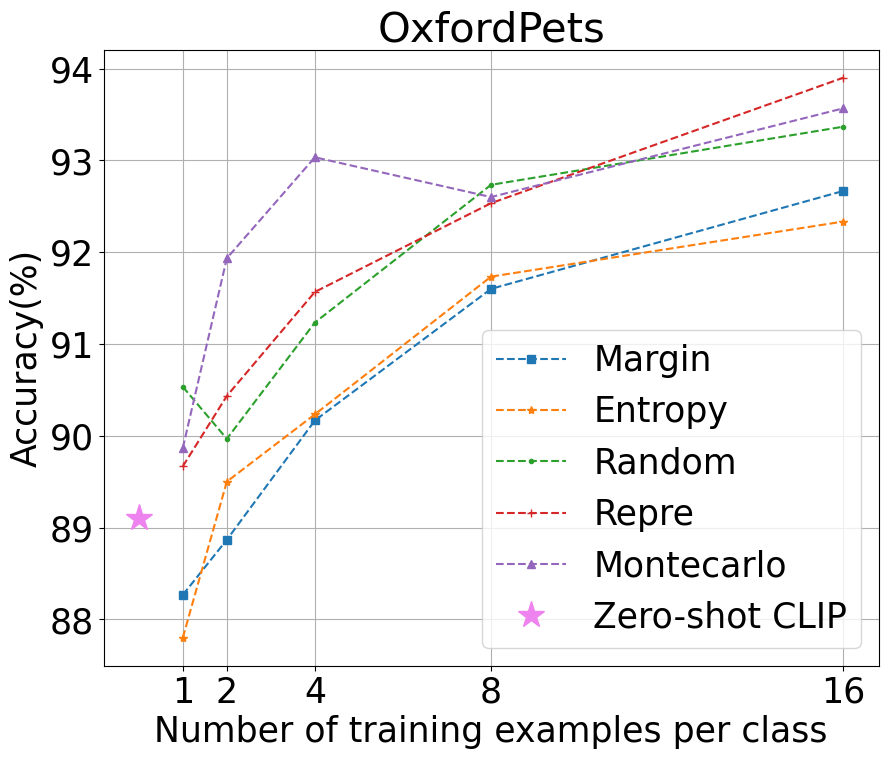} \\
\includegraphics[width=0.23\textwidth]{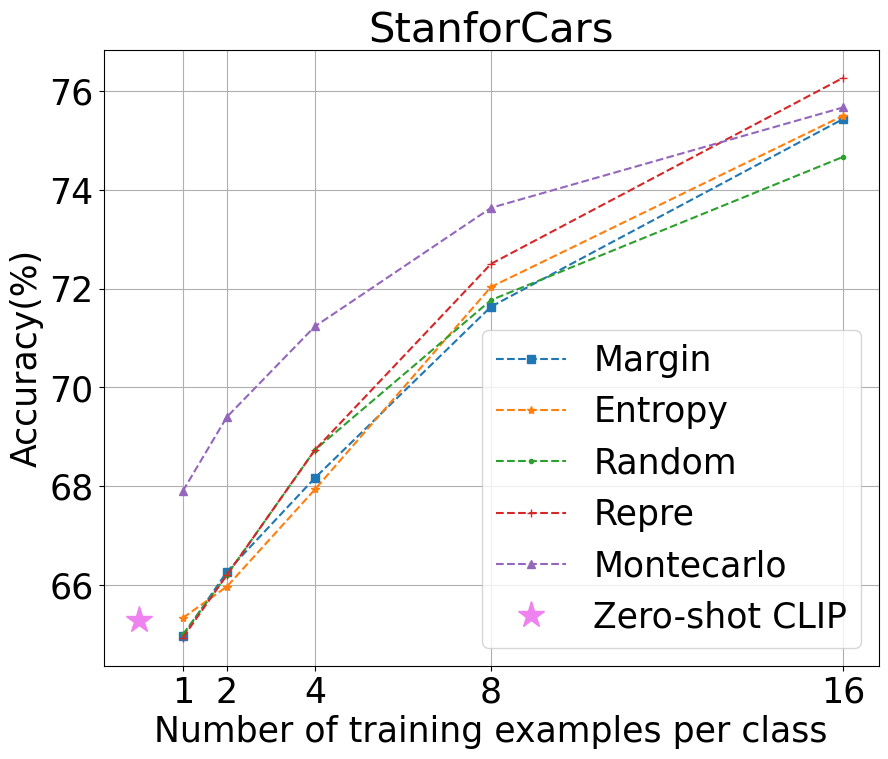} &
\includegraphics[width=0.23\textwidth]{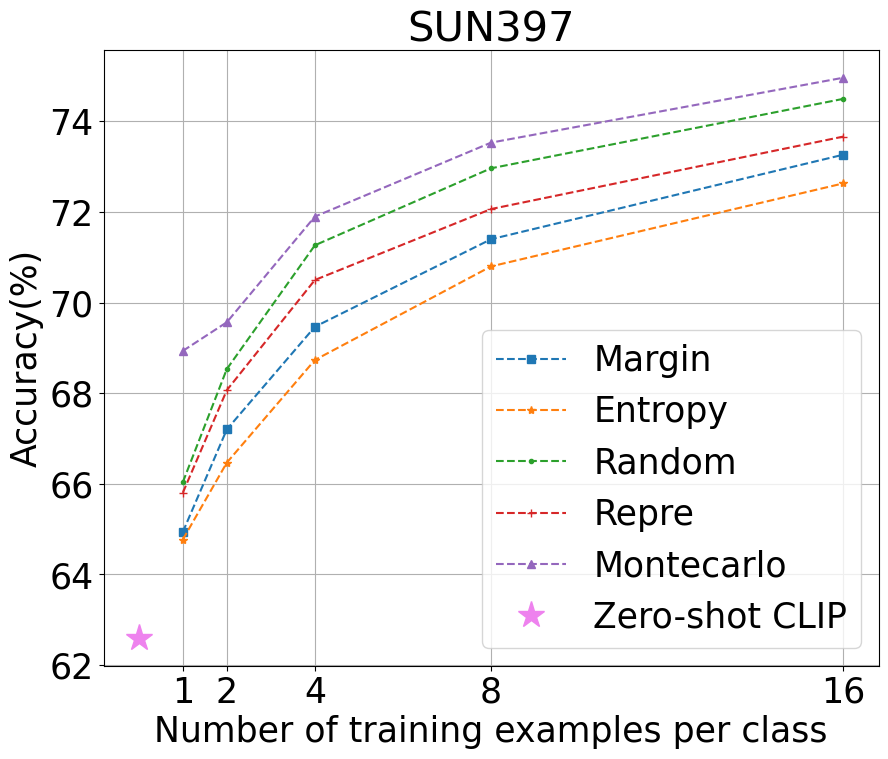} &
\includegraphics[width=0.23\textwidth]{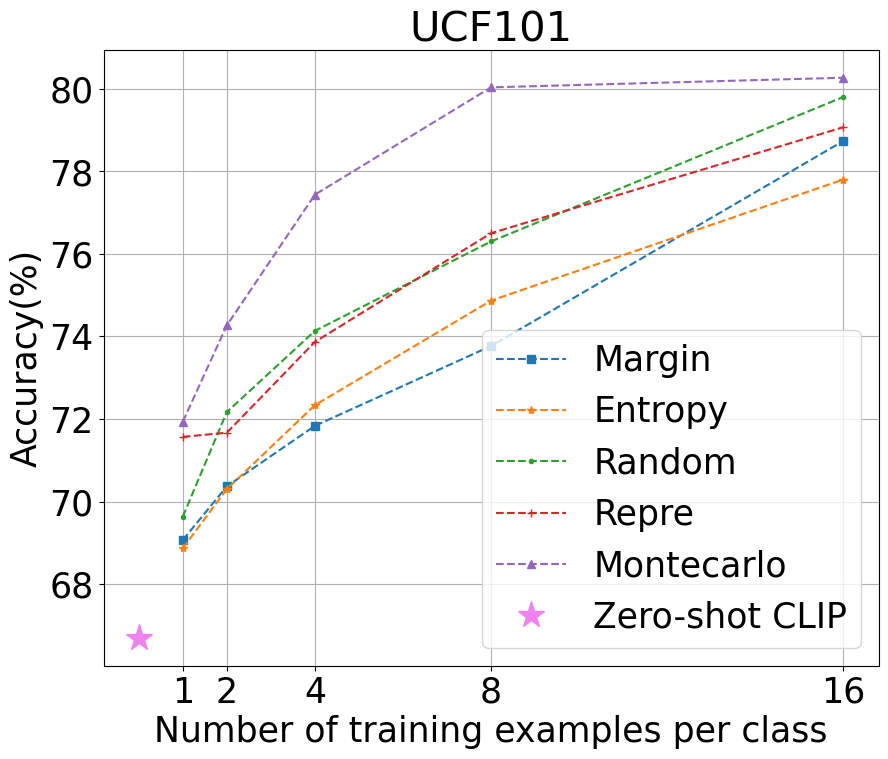} \\
\end{tabular}
\caption{Main results of few-shot learning on the 11 datasets in CoOP. The results showcase performance variability across different domains and illustrate the method's generality and adaptability.}
       \Description{Illustration of Main results}
\label{fig:coop}
\end{figure*}

\begin{figure*}[htbp]
\centering
\begin{tabular}{ccc}
\includegraphics[width=0.23\textwidth]{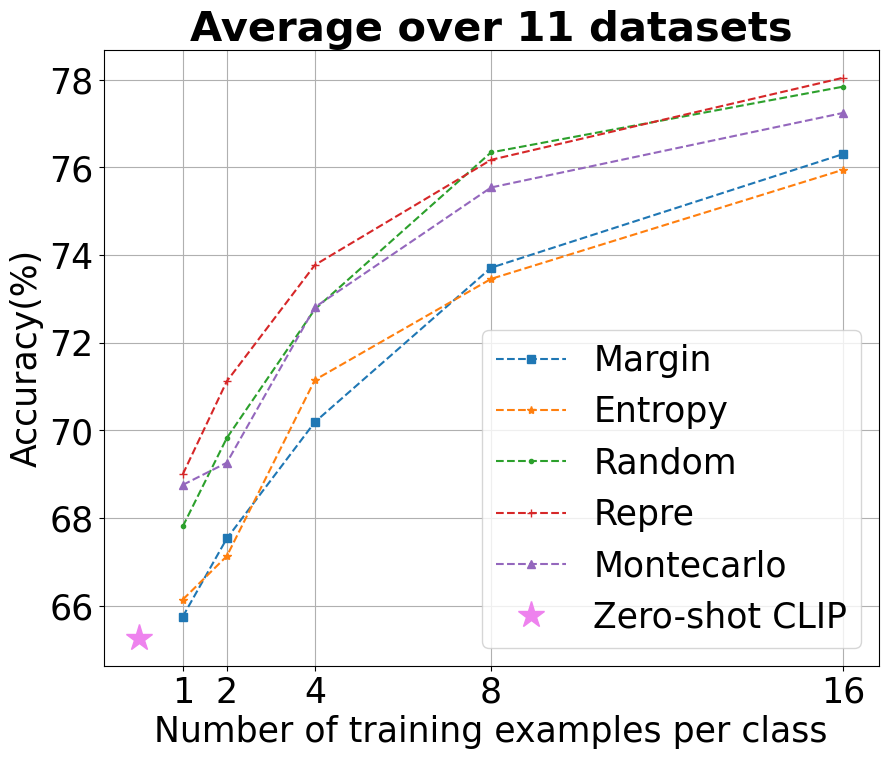} &
\includegraphics[width=0.23\textwidth]{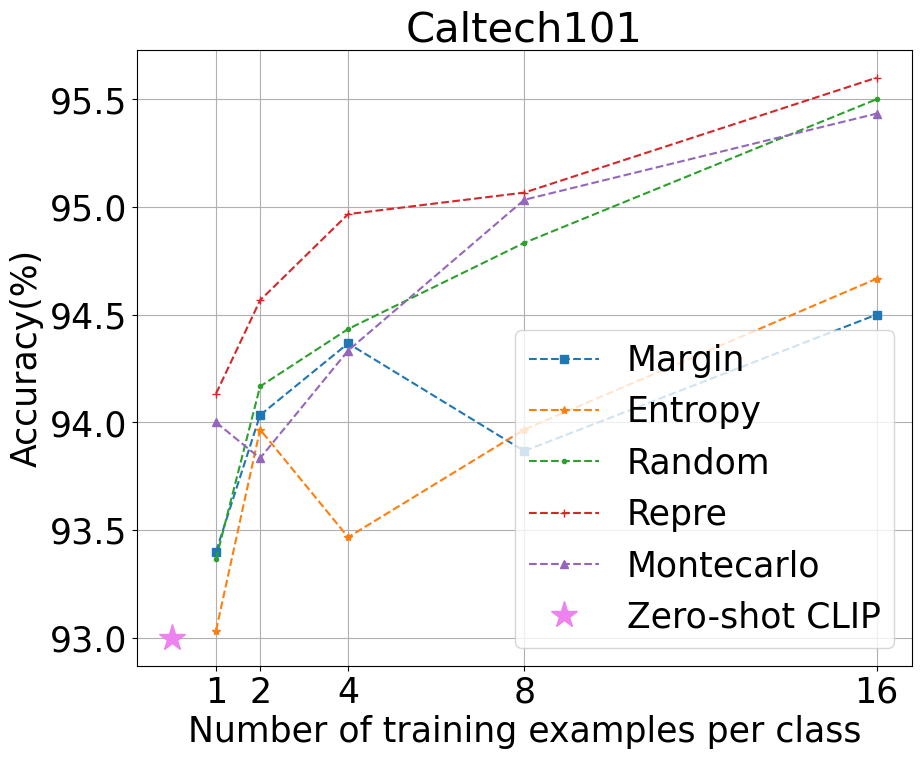} &
\includegraphics[width=0.23\textwidth]{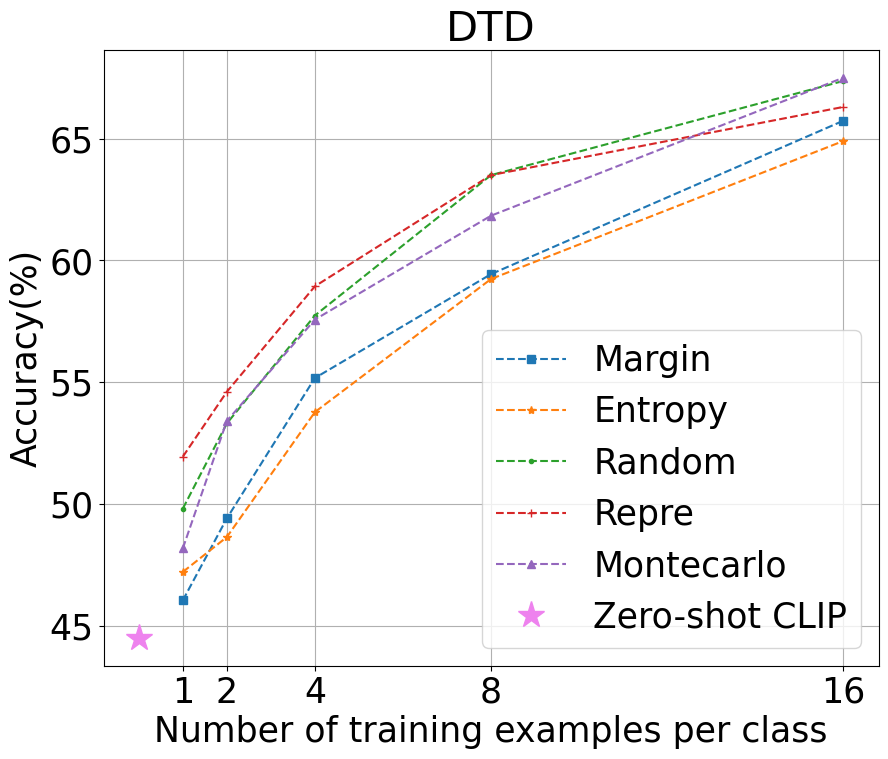} \\
\includegraphics[width=0.23\textwidth]{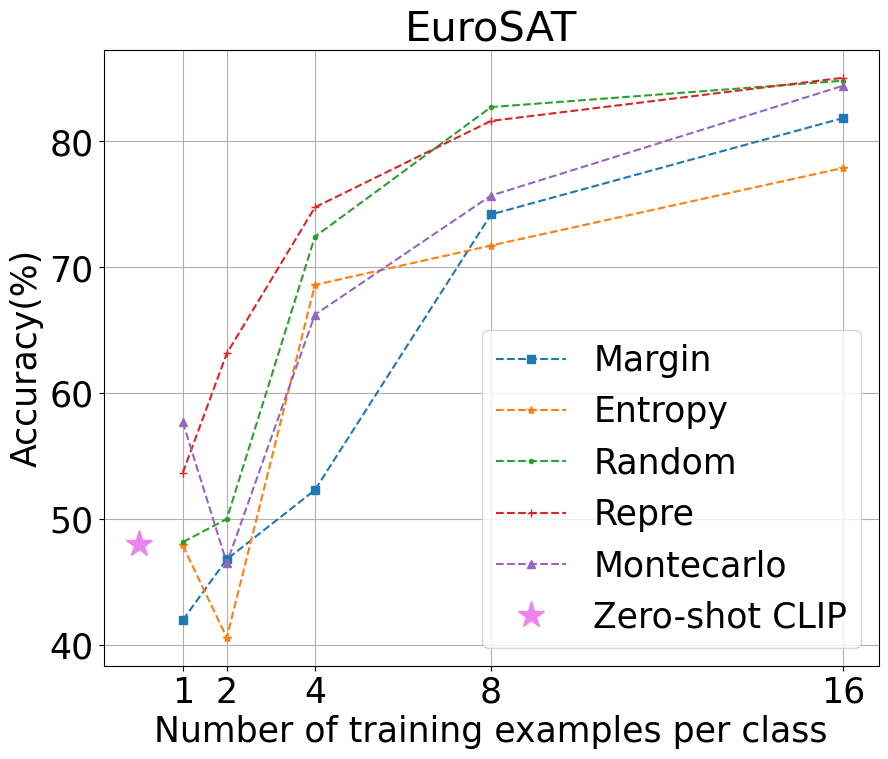} &
\includegraphics[width=0.23\textwidth]{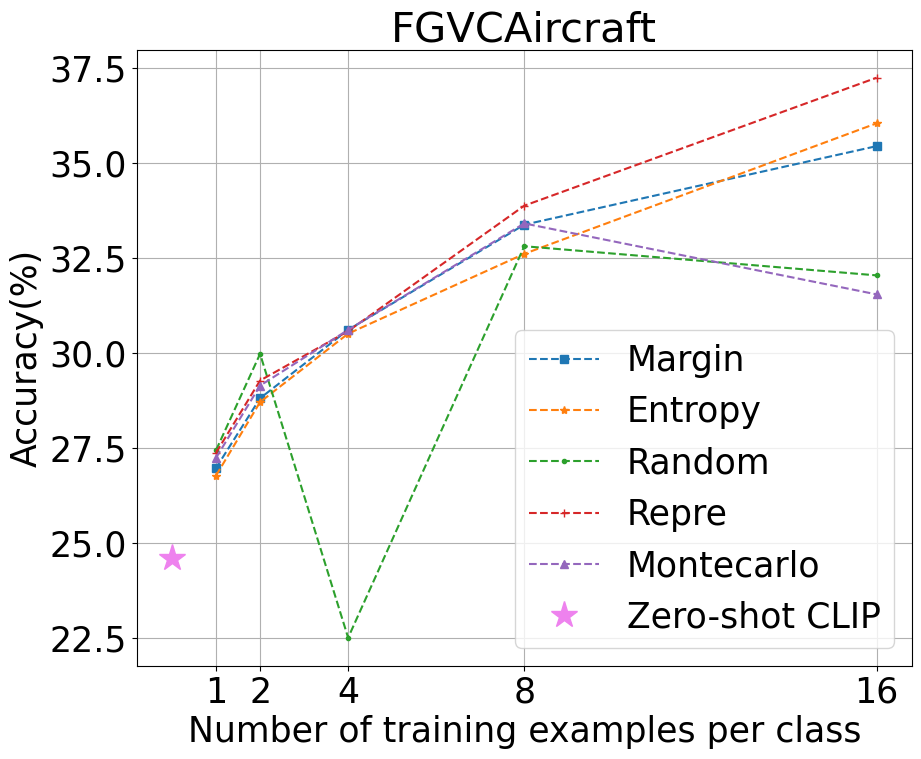} &
\includegraphics[width=0.23\textwidth]{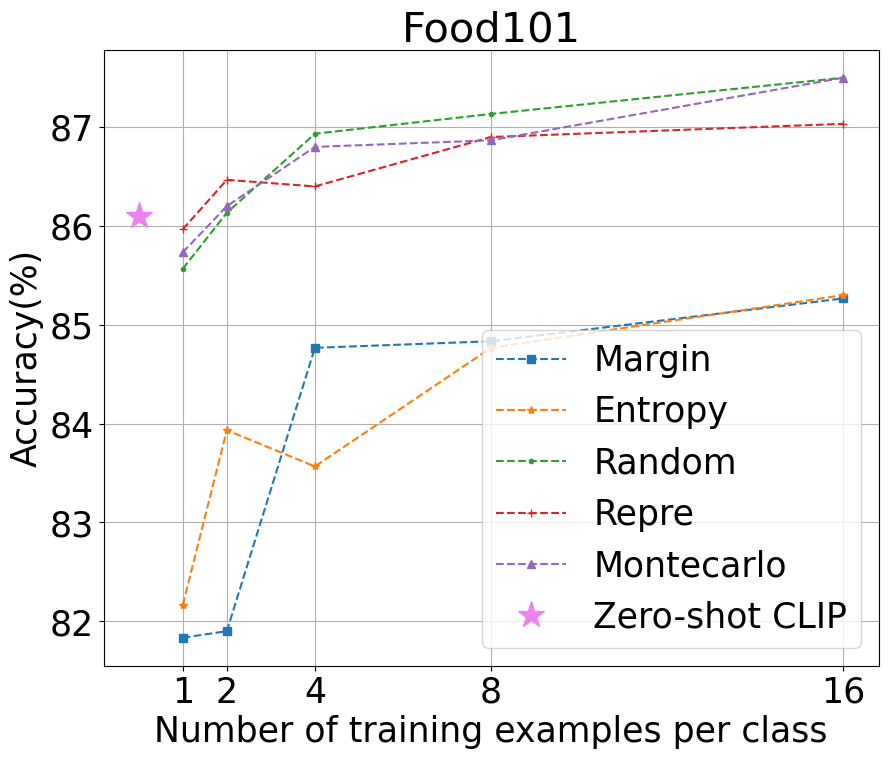} \\
\includegraphics[width=0.23\textwidth]{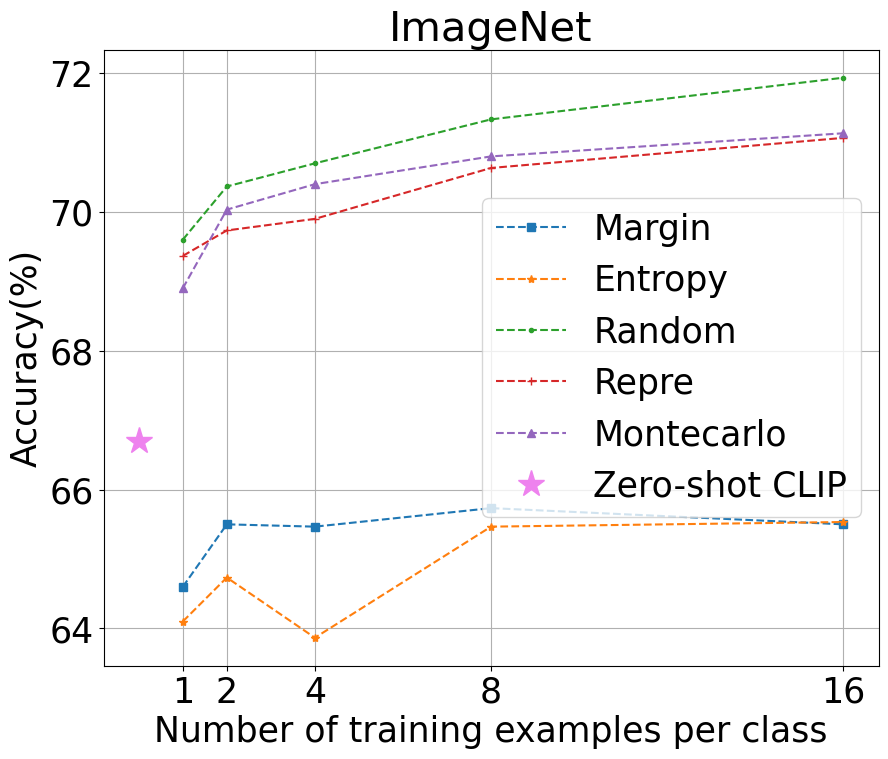} &
\includegraphics[width=0.23\textwidth]{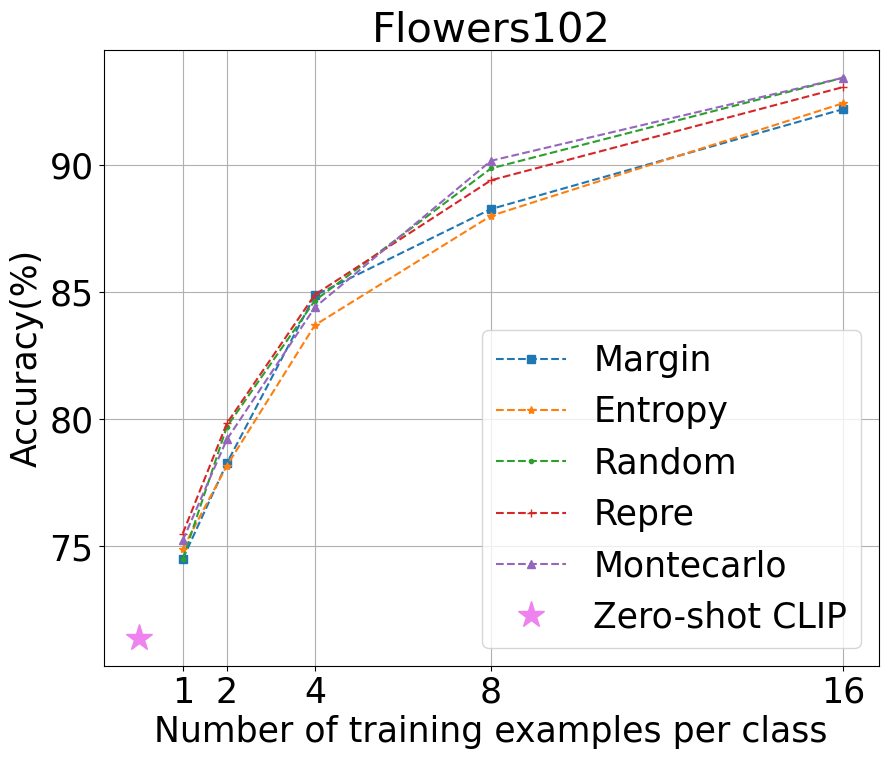} &
\includegraphics[width=0.23\textwidth]{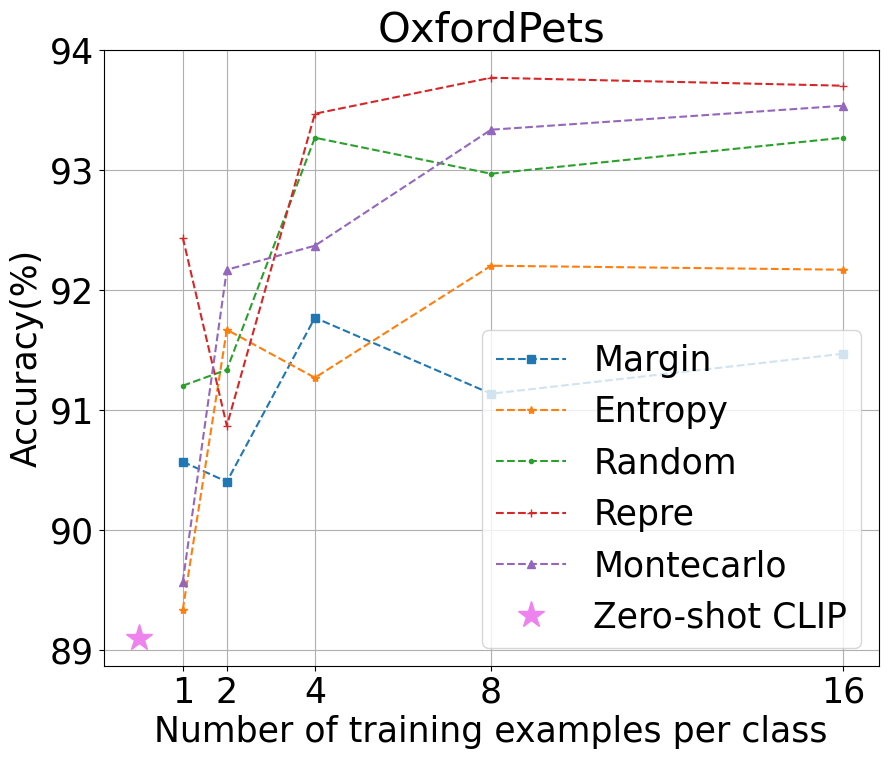} \\
\includegraphics[width=0.23\textwidth]{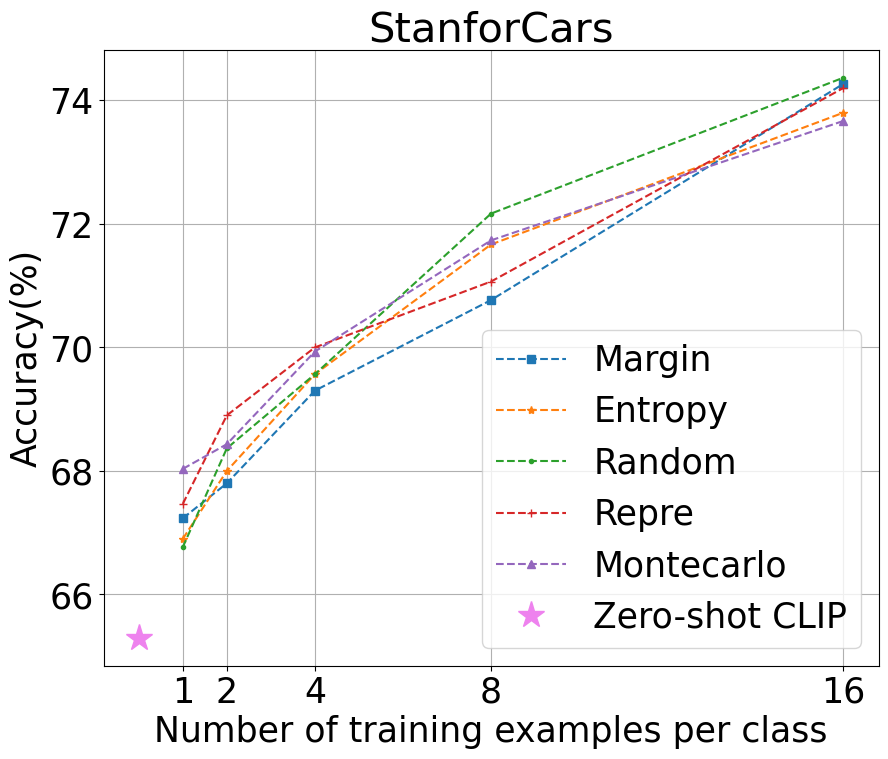} &
\includegraphics[width=0.23\textwidth]{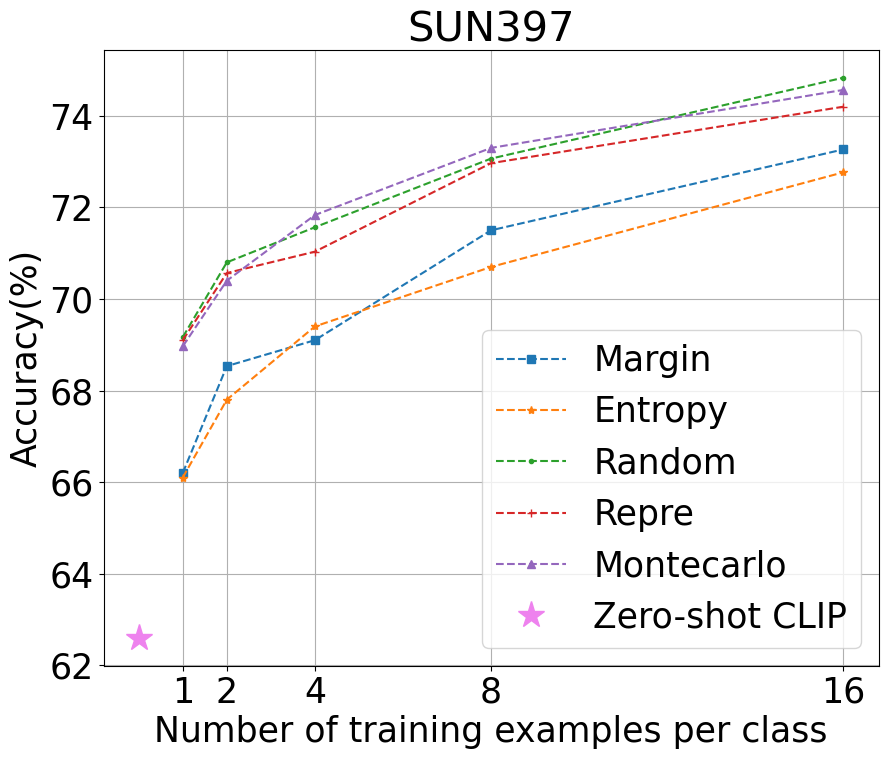} &
\includegraphics[width=0.23\textwidth]{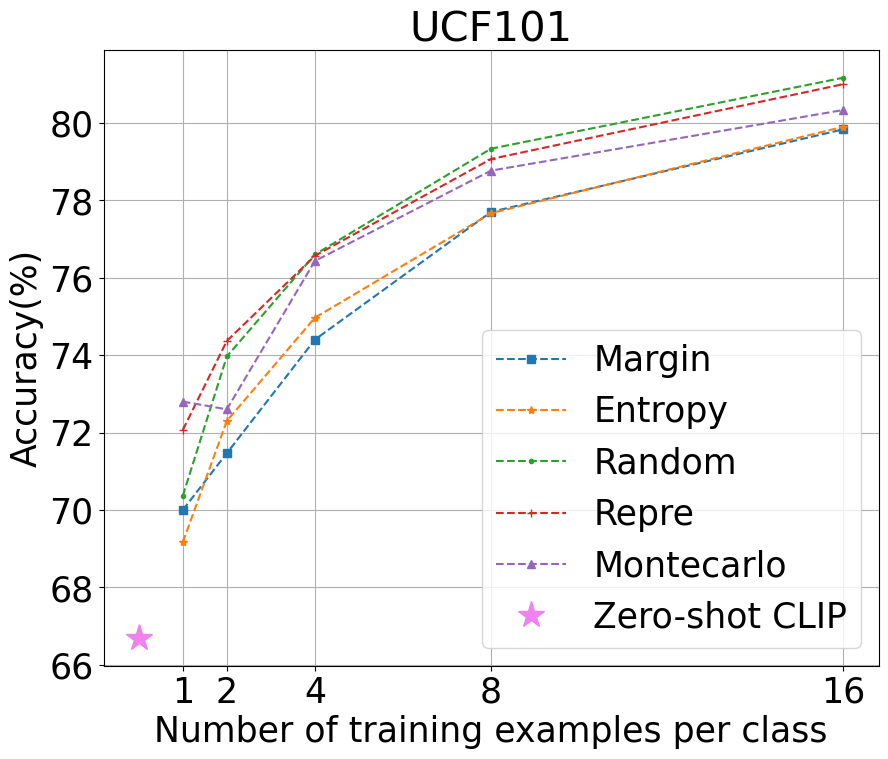} \\
\end{tabular}
\caption{Main results of few-shot learning on the 11 datasets in MaPle. The visualizations depict performance across various domains, highlighting the effectiveness and adaptability of the MaPle method.}
       \Description{Illustration of MaPle results}
\label{fig:maple}
\end{figure*}


\section{Experiment}
In this section, we will present the experimental setup, datasets, implementation details as well as the experimental results in detail.
\subsection{Datasets}
For our few-shot settings, we use the 11 image recognition datasets as in~\cite{CoOp}, which cover a diverse set of recognition tasks to construct a comprehensive benchmark. Specifically, the benchmark includes ImageNet~\cite{imagenet} and Caltech101~\cite{caltech101} for classification on generic objects; OxfordPets~\cite{oxfordpets}, StanfordCars~\cite{StanfordCars}, Flowers102~\cite{oxford_flowers}, Food101~\cite{food} and FGVCAircraft~\cite{aircraft} for fine-grained classification; SUN397~\cite{sun} for scene recognition; UCF101~\cite{ucf101} for action recognition; DTD~\cite{dtd} for texture classification; and finally EuroSAT~\cite{eurosat} for satellite imagery recognition.

\subsection{Setting}

\paragraph{Few-shot setting}
In the few-shot learning scenario, we evaluate the effectiveness of training with 1, 2, 4, 8, and 16-shot datasets, testing against the comprehensive test sets. This methodology aligns with the approaches outlined in~\cite{CoOp, Maple, zhou2022conditional}.

\paragraph{Few-shot Frameworks}\label{subsec:fs-method}
We perform the data selection techniques on three commonly used few-shot learners: \textbf{Linear probe} CLIP~\cite{clip-adapter} trains an additional linear classifier on top of its visual encoder and follows a few-shot training manner. \textbf{MaPLe}~\cite{Maple} jointly optimizes visual prompt and text prompt in the CLIP model, which leads to better performance. \textbf{CoOp}~\cite{CoOp} fine-tunes CLIP for few-shot image classification by optimizing the continuous set of prompt vectors at its language branch.

\subsection{Baselines}
Previous work such as~\cite{CoOp,Maple,zhou2022conditional} uses a random selection strategy which is denoted as \textbf{Random}. In our work, we adopt Random as a baseline. As described in Section~\ref{subsec:active_learning}, we use CLIP zero-shot to compute the logits for each class and select examples with minimal difference between the first and second largest logits, which is termed as \textbf{Margin}. Another prevalent method in AL involves using entropy to identify examples with the highest uncertainty, as detailed in Section~\ref{subsec:active_learning}. This approach is termed as \textbf{Entropy}. We adopt both Margin and Entropy as strong data selection baselines. Each baseline methods will be integrated with the three few-shot learning frameworks detailed in~\ref{subsec:fs-method} to obtain the outcomes. We also report the result of \textbf{zero-shot CLIP} as a weak baseline.


\subsection{Implementation Details}
Our proposed Gaussian Monte Carlo selection strategy, as detailed in Section~\ref{subsec:Montecarlo}, henceforth referred to as \textbf{Montecarlo}, incorporates Gaussian noise characterized by a mean of 0 and a standard deviation of 0.1. Additionally, another methodology that leverages representativeness for the selection of few-shot examples, as detailed in Section~\ref{subsec:repre}, is designated as \textbf{Repre}. We compute distance between examples using cosine similarity.

For all experiments, we fine-tuned the CoOp model~\cite{CoOp}, MaPLe model~\cite{Maple} and linear probe CLIP model~\cite{CLIP} with backbone ViT-B/16 CLIP model where $d_l = 512$, $d_v = 768$ and $d_vl = 512$. MaPLe is trained for 5 epochs with a batch size of 4 and a learning rate of 0.002 via SGD optimizer on a single NVIDIA A100 GPU. CoOp is trained for 10 epochs with a batch size of 32 and a learning rate of 0.002 via SGD optimizer on a single NVIDIA A100 GPU. We use a photo of a <category> as a template to acquire pre-trained CLIP word embeddings.

For each experimental set, we conduct three different random seeds to mitigate the impact of randomness, and then report the averaged results. For Margin and Entropy, we apply CLIP zero-shot model with backbone ViT-B/16 to compute the logits. 

\subsection{Main Results}

\subsubsection{Overall Performance}\label{subsec: overall_performance}

Figs.~\ref{fig:coop},~\ref{fig:maple} and 4 (for brevity, see in Appendix \ref{appendix:linear}) show the performance of different data selection techniques combined with CoOp, MaPLe and Linear probe respectively. Experimental results are evaluated over 11 distinct datasets, demonstrated by different subplots. Each subplot shows the accuracy (y-axis) against the number of training examples per class (x-axis), varying from 1 to 16 examples.

It can be observed that in nearly all cases, increasing the number of training examples per class leads to performance uplift, which aligns with anticipations, as the provision of additional data enables a more substantial delineation of decision boundaries.

Significantly, in almost all scenarios, the proposed methods, Montecarlo and Repre, demonstrate a very competitive performance. This is evidenced by the significant improvement they achieve compared to baseline methods such as Random, Entropy, and Margin. This improvement can be generalized across different datasets and different few-shot learners, as shown by the results.

Based on the experimental outcomes depicted in the graphs, it appears that the performance of AL methods, particularly Margin and Entropy, is not as advantageous as one might expect when compared to random selection in few-shot learning scenarios. This observation resonates with the findings reported in ~\citet{Active}, which asserts that merely incorporating active learning strategies into few-shot contexts does not yield the expected improvements.


\subsubsection{Results using CoOp}

Fig.~\ref{fig:coop} shows the performance of different data selection techniques combined with CoOp learner.

In this case, the Montecarlo method demonstrates the best performance across different datasets. To be concrete, the averaged performance on 11 datasets indicates that the Montecarlo method demonstrates a robust performance trajectory, which suggests that it benefits from its inherent ``unfamilarity'' to explore the data space comprehensively. Furthermore, the Montecarlo method shows a commendable scalability with an increase in number of training examples, evidenced by the consistent upward trend in accuracy as the number of training examples per class grows.

Interestingly, in the context of using CoOp, it is observed that Random selection emerges as the second-best method, with Repre demonstrating comparable effectiveness. However, the AL-based methods, Entropy and Margin, appear to significantly underperform in relation to both Random and Repre. This suggests that in the specific framework of CoOp, the traditional advantages of active learning strategies like Entropy and Margin are not effective.

For datasets such as Caltech101, Flowers102, and Oxford Pets, all methodologies swiftly achieve high levels of accuracy with only a handful of examples. This could suggest that the features within these datasets are less intricate, or that the classes are more readily distinguishable from one another. Within these particular datasets, the Montecarlo method displays a distinct advantage, achieving the highest accuracy levels when the training examples are extremely limited, notably when there is just one or two examples per class.

For datasets like ImageNet and SUN397, the accuracy tends to be relatively low, and the benefits of adding more training examples per class are limited, as demonstrated by the plateau in the curve. This suggests that datasets with a high degree of complexity or diversity, considering their wide range of classes and domains, gain less from the inclusion of additional few-shot examples. Nonetheless, in these cases, the Montecarlo method still manages to secure the best or competitive results. Specifically within the ImageNet dataset, the AL methods, Margin and Entropy, perform significantly worse than Random, Montecarlo, and Repre, even falling below zero-shot performance levels. This further illustrates that merely implementing AL methods in few-shot scenarios has limited effectiveness.

Overall, when employing CoOp, the introduced Montecarlo and Repre methods can attain the best or competitive results compared to Random, while notably surpassing the performance of AL-based methods like Entropy and Margin.

\subsubsection{Results using MaPLe and Linear Probe}

Figures~\ref{fig:maple} and ~\ref{fig:linear} (see Appendix\ref{appendix:linear}) display the efficacy of various data selection strategies when used alongside with MaPLe and Linear Probe, respectively. The results here slightly diverge from those observed with CoOp. Specifically, for both MaPLe and Linear Probe frameworks, Repre emerges as the top-performing method, followed by Random, with Montecarlo showing competitiveness comparable to Random. In contrast, similar to the results of CoOp, Margin, and Entropy methods significantly underperform those of Random, Repre, and Montecarlo, underscoring once more the limitations of AL-based methods in a few-shot scenario.

Notably, within the Linear Probe framework, when the number of training examples is extremely limited (i.e., 1 or 2 examples per class), Repre stands out as the sole method capable of matching or surpassing the accuracy of CLIP in a zero-shot setting. This further highlights Repre's superior performance in scenarios where the few-shot framework is considered ``weak.'' Typically, Linear Probe is regarded as a less capable few-shot framework as compared to those few-shot prompt learners such as MaPLe and CoOp. The remarkable efficiency of Repre suggests its unique capability to leverage minimal data for maximum performance. This efficiency may stem from representativeness of the selected data, enabling it to extract and utilize critical features from very few examples effectively. Such an attribute is crucial in few-shot learning environments, where the ability to learn from minimal information is paramount.

\section{Analysis}

\subsection{Why AL-based methods fail?}

Consistently poorer performance of two AL-based methods across various datasets and few-shot frameworks show the ineffectiveness of AL strategies in the context of few-shot learning. We presume the subpar performance stems from the following two factors: 

\textbf{AL methods highly depend on initial model performance:} AL methods typically rely on the predictions of the current model to determine which data points would be most valuable. In few-shot settings, the initial model is often worse, leading to poor initial performance. This can result in a skewed understanding of which examples are truly informative, as the model's predictions might not accurately reflect the underlying distribution of the data. In contrast, methods like Montecarlo and Repre reduce dependence on the initial model's prediction accuracy by relying on the characteristics of the data itself, for example, computing distance between examples in embedding space.

\textbf{Noise Sensitivity:} Entropy-based methods (i.e., Margin and Entropy) select examples that the model is most uncertain about. In few-shot settings, where models are more prone to overfitting or underfitting due to the limited data, the model's uncertainty (measured by entropy) might not accurately represent the true informativeness of an example. In practice, it's often noted that the logit values across different classes are remarkably close, with differences extending to nearly three decimal places. This observation points to a limited distinguishing capability in the initial model. In such case, the model's uncertainty might be more influenced by noise in the data, leading to the selection of outliers or anomalous examples that do not provide generalizable insights for improving the model. 

\subsection{What causes the varied behaviors of Montecarlo within MaPLe frameworks?}

When employing CoOp, Montecarlo demonstrates a clear advantage, whereas with MaPLe and Linear Probe, its effectiveness diminishes. This variance in performance is likely due to the unique characteristics and features of the visual branches within different datasets, or more specifically, the generality of these datasets.


To assess the datasets' generality and examine its relationship with the effectiveness of the Montecarlo method when integrated with the MaPLe framework, we calculate the cosine similarity across all examples within each dataset. We perform an analysis by classifying datasets into ``Good'' or ``Bad'' categories, contingent on their performance outcomes with Montecarlo, particularly within the framework of MaPLe.

Table~\ref{tab:ablation} lists datasets along with their corresponding cosine similarity values. The cosine similarity here likely represents how similar examples within each dataset are to each other, which can be interpreted as a measure of the dataset's generality. Higher cosine similarity values suggest that examples within the dataset are more similar to each other, indicating a more specific or less diverse dataset.

The table divides datasets into two groups based on their performance with the Montecarlo method in the context of the MaPLe framework. The ``Good'' group contains datasets where Montecarlo shows improved performance, while the ``Bad'' group lists datasets where Montecarlo results in comparatively lower performance. It's noted that the average cosine similarity for the ``Good'' group is higher than that for the ``Bad'' group, suggesting that Montecarlo tends to perform better on datasets with lower generality (smaller similarity among examples).

This finding suggest that the Montecarlo method's effectiveness, especially when combined with the MaPLe framework, is sensitive to the degree of generality within the dataset. Datasets with higher cosine similarity values, indicating a more uniform or less diverse set of examples, seem to benefit more from Montecarlo's instance selection strategy. This could be because Montecarlo, by focusing on selecting unfamiliar examples, may be more effective when the overall dataset exhibits higher internal similarity, allowing the method to identify and leverage the nuances among examples more effectively.

Conversely, datasets with higher generality (or higher diversity among examples), as seen in the ``Bad'' group, might pose challenges for Montecarlo's selection strategy. In these cases, the intrinsic diversity within the dataset could make it more difficult for Montecarlo to identify examples that are significantly more informative or unfamiliar compared to the rest, thereby diminishing the potential benefits of this selection approach.

In summary, the effectiveness of the Gaussian Monte Carlo method in selecting few-shot examples, especially when used alongside the MaPLe framework, seems to be influenced by the dataset's generality. The impact of the internal characteristics of image data likely stems from MaPLe's approach of simultaneously tuning both the visual and text prompts, whereas CoOp only adjusts the text prompt. Montecarlo shows improved performance with datasets characterized by higher cosine similarity, which suggests a need for further exploration into how instance selection strategies can be optimized based on the specific attributes and diversity of the datasets being used for few-shot learning.

\begin{table}[htbp]
  \caption{Ablation Study: Using cosine similarity to evaluate the degree of generality of the dataset.}
  \label{tab:ablation}
  \centering 
  \scalebox{0.9}{
    \begin{tabular}{cc}
      \hline
      \textbf{Dataset} & \textbf{Cosimilarity} \\
      \hline
      \textbf{Good} \\
      \hline
      Caltech101 & 65.18   \\
      Dtd & 81.68  \\
      Eurosat & 89.685   \\
      Fgvc aircraft & 66.3 \\
      Oxford pets & 91.5 \\
      Ucf101 & 60.406 \\
      Mean & 75.69 \\
      \hline
      \textbf{Bad} \\
      \hline
      Oxford Flowers & 88.625  \\
      Stanford cars & 65.375  \\
      Food101 & 73.125  \\
      Sun397 & 58.2185 \\
      Mean & 70.24 \\
      \hline
    \end{tabular}
  }
\end{table}

\subsection{What makes good few-shot examples?}

Drawing from the experimental observations, we infer that effective few-shot examples in VL models are influenced by several factors:

\textbf{Decoupling from model prediction}: The subperformance of uncertainty-based methods, such as Margin and Entropy, suggests that in selecting few-shot examples, it is crucial to decouple the process from model predictions. This is due to the indistinguishable ability of VL models in few-shot scenarios. It is more effective to rely on the features or characteristics of the data itself.

\textbf{Unfamiliarity and Diversity}: Good few-shot examples are those that introduce the most unfamiliar or diverse concepts to the model relative to its pre-trained knowledge. This criterion is crucial because models learn more effectively when exposed to new, informative data that fills in the gaps in their existing knowledge. The Montecarlo method leverages this principle by selecting examples based on their distance from modified versions with added Gaussian noise, aiming to identify instances that are less familiar to the model and, hence, more valuable for learning.

\textbf{Representativeness}: Examples that are highly representative of their respective classes or categories tend to make good few-shot learning instances. The REPRE method focuses on selecting examples that are closest to the centroid of their class in the embedding space, ensuring that the chosen few-shot examples provide a comprehensive overview of the data's diversity. This approach helps in selecting instances that encapsulate the essential features of their class, making them particularly effective for few-shot learning.

\textbf{Synergy Between Selection Method and Few-Shot Frameworks:} A critical aspect of determining good few-shot examples involves considering the synergy between the selection method used and the specific few-shot learning frameworks being applied. Different frameworks, such as CoOp, MaPLe, or Linear Probe, may have varying strengths and weaknesses, which can be complemented or exacerbated by the choice of example selection method. For instance, the Montecarlo method may perform exceptionally well with a framework like CoOp by leveraging its ability to identify and utilize unfamiliar examples, thereby enhancing the model's learning efficiency. On the other hand, the REPRE method might show superior performance with frameworks like MaPLe or Linear Probe by focusing on the most emblematic examples of the data distribution, thereby ensuring that the few-shot examples are highly informative and characteristic of their respective classes.
\section{Conclusion}
In this work, we explore the novel challenge of selecting training instances for few-shot image recognition using pre-trained vision-language models. We demonstrate that random sampling introduces considerable variability and often results in subpar performance. Surprisingly, we also find that Active Learning methods, such as Margin and Entropy, consistently underperform compared to random selection. To address this issue, we introduce two new strategies --- Gaussian Monte Carlo and Representativeness-based selection approach. Our results show that new methods not only surpasses the effectiveness of random sampling but also outperforms other AL-based data selection methods. We also conduct a comprehensive, in-depth analysis of how these new approaches impact few-shot scenarios.



\bibliographystyle{ACM-Reference-Format}
\bibliography{sample-base}

\clearpage  

\appendix
\section{Results of Linear Probe}\label{appendix:linear}
\noindent
\begin{minipage}{\textwidth}
\centering
\begin{tabular}{ccc}
\includegraphics[width=0.3\textwidth]{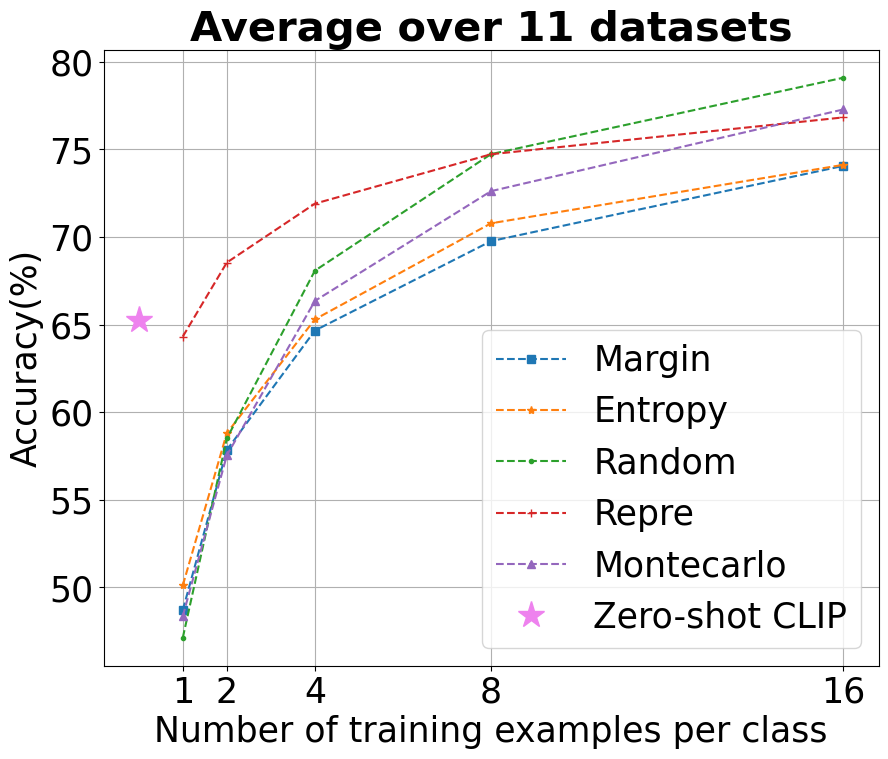} &
\includegraphics[width=0.3\textwidth]{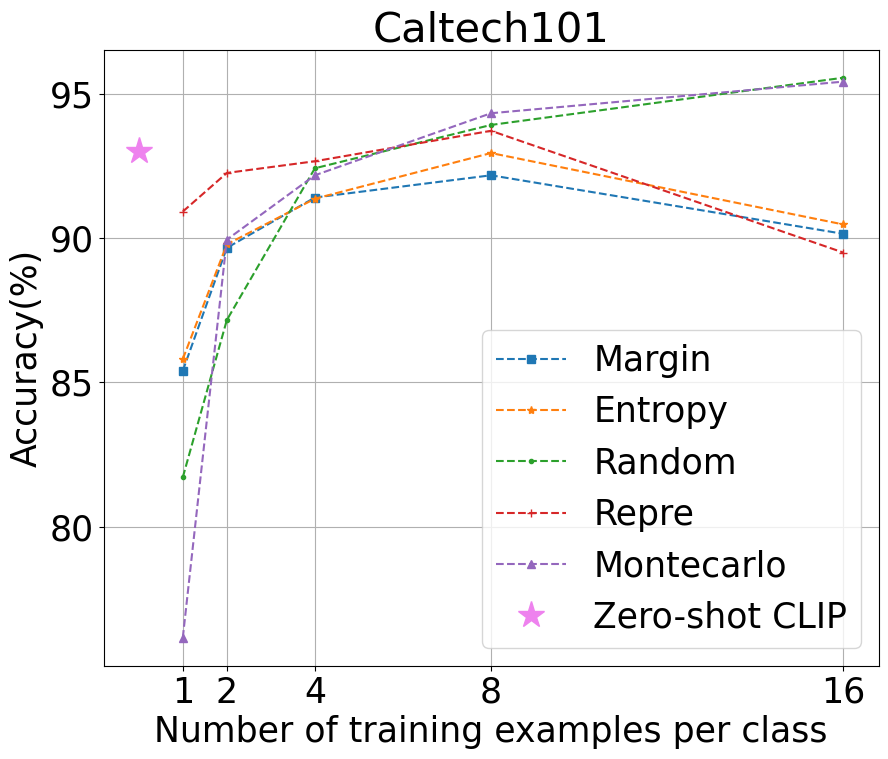} &
\includegraphics[width=0.3\textwidth]{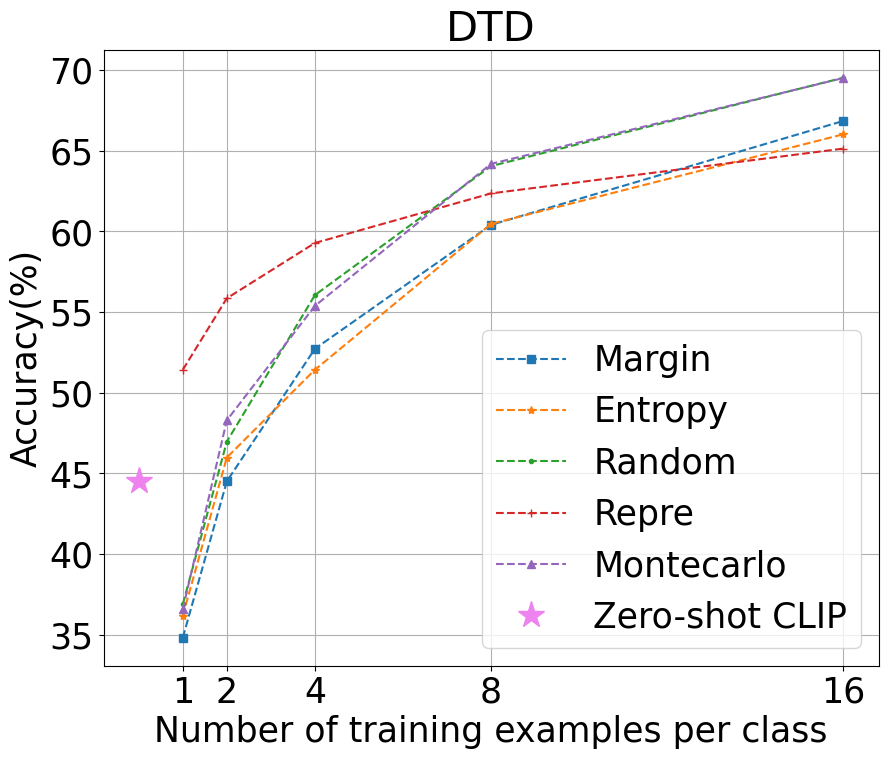} \\
\includegraphics[width=0.3\textwidth]{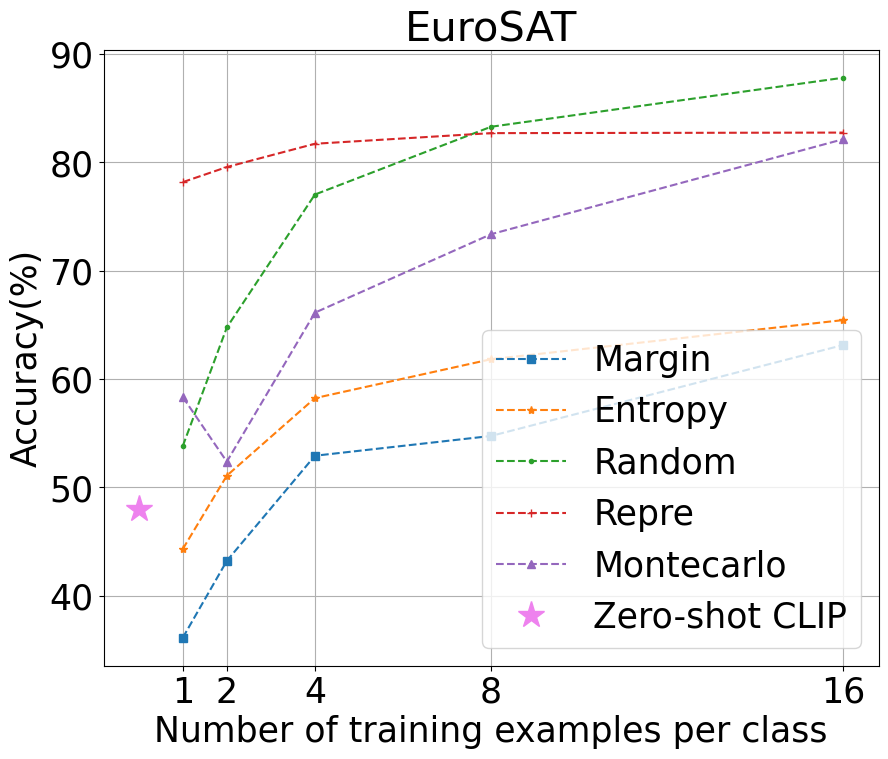} &
\includegraphics[width=0.3\textwidth]{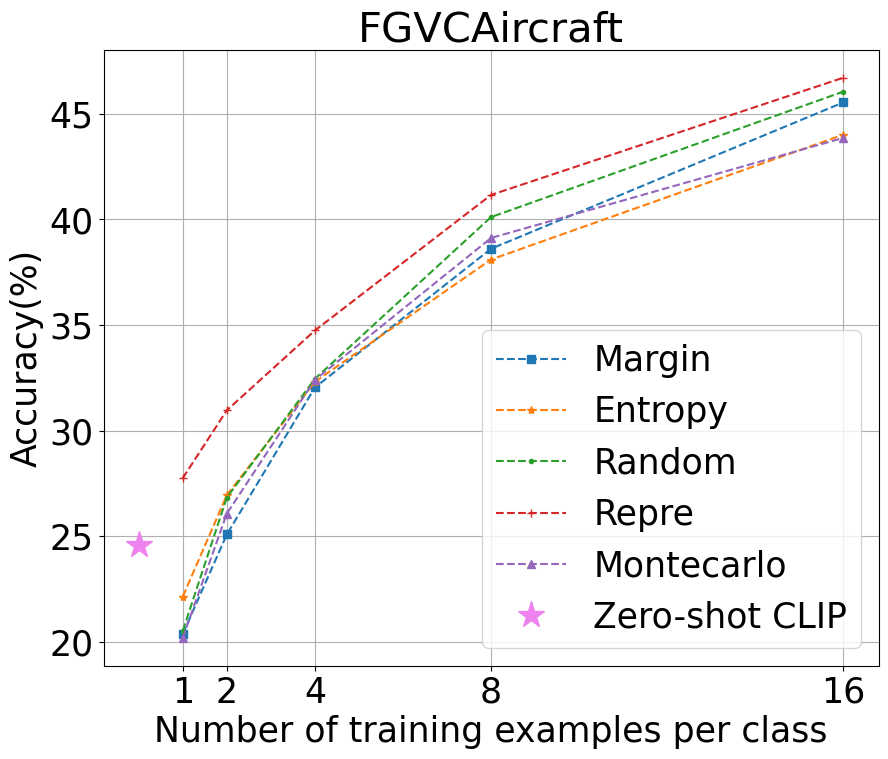} &
\includegraphics[width=0.3\textwidth]{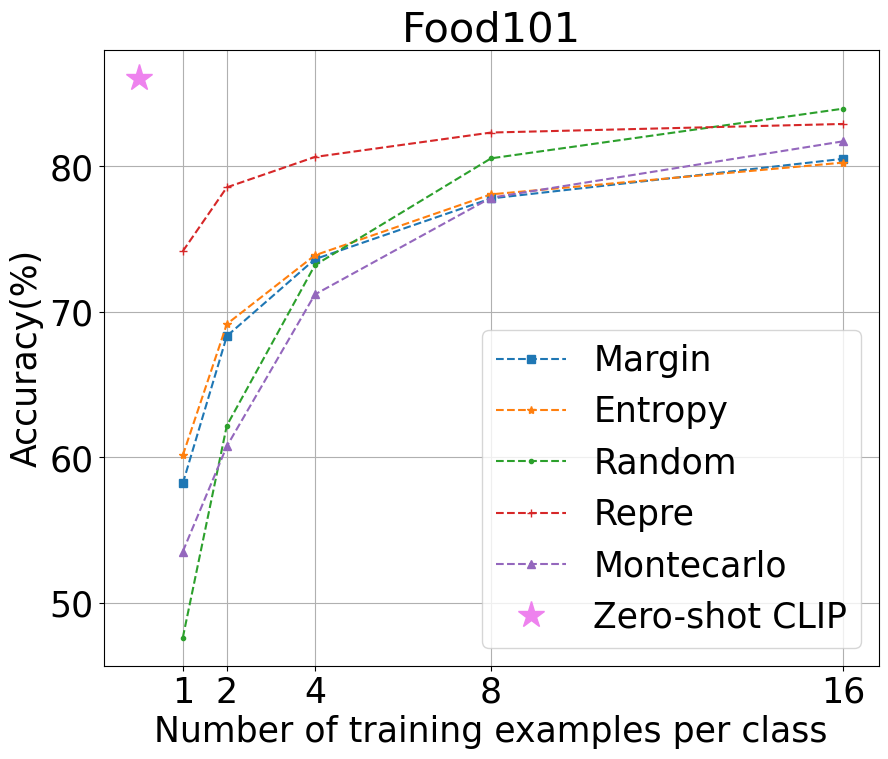} \\
\includegraphics[width=0.3\textwidth]{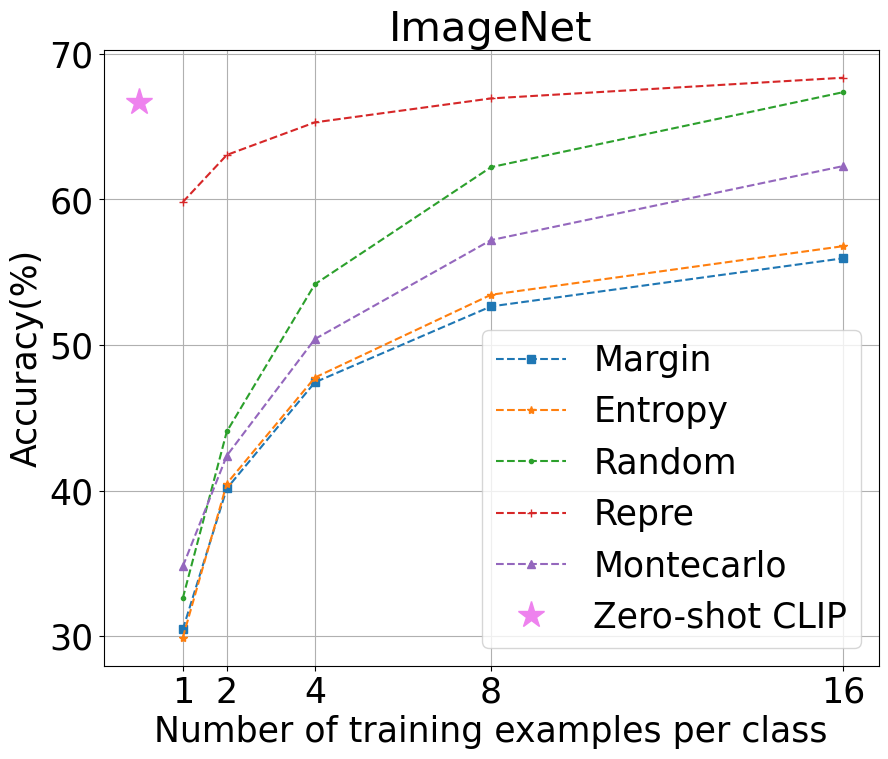} &
\includegraphics[width=0.3\textwidth]{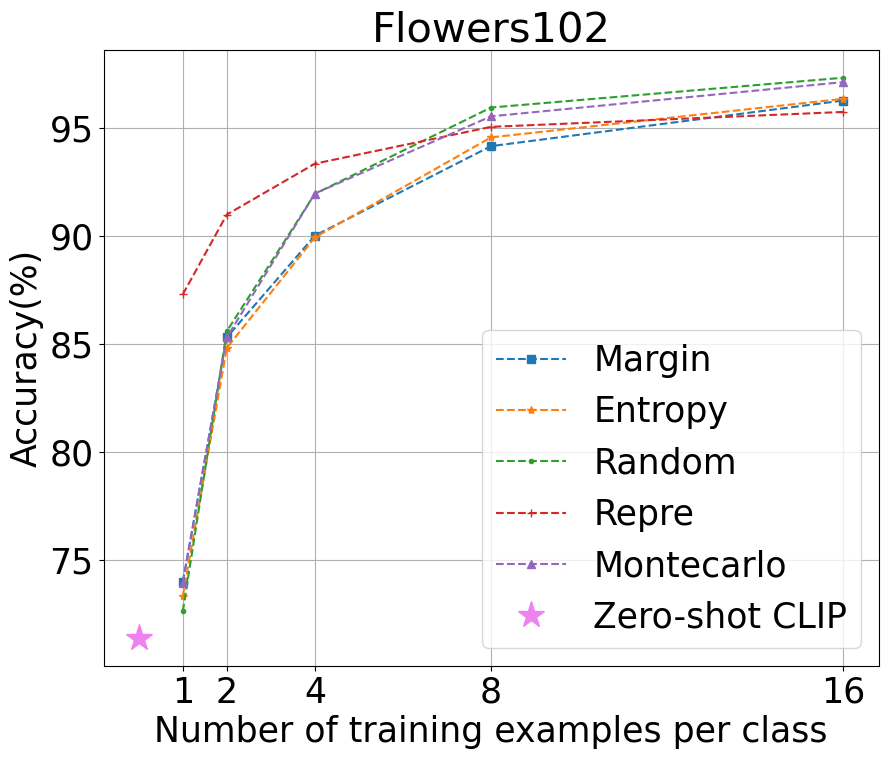} &
\includegraphics[width=0.3\textwidth]{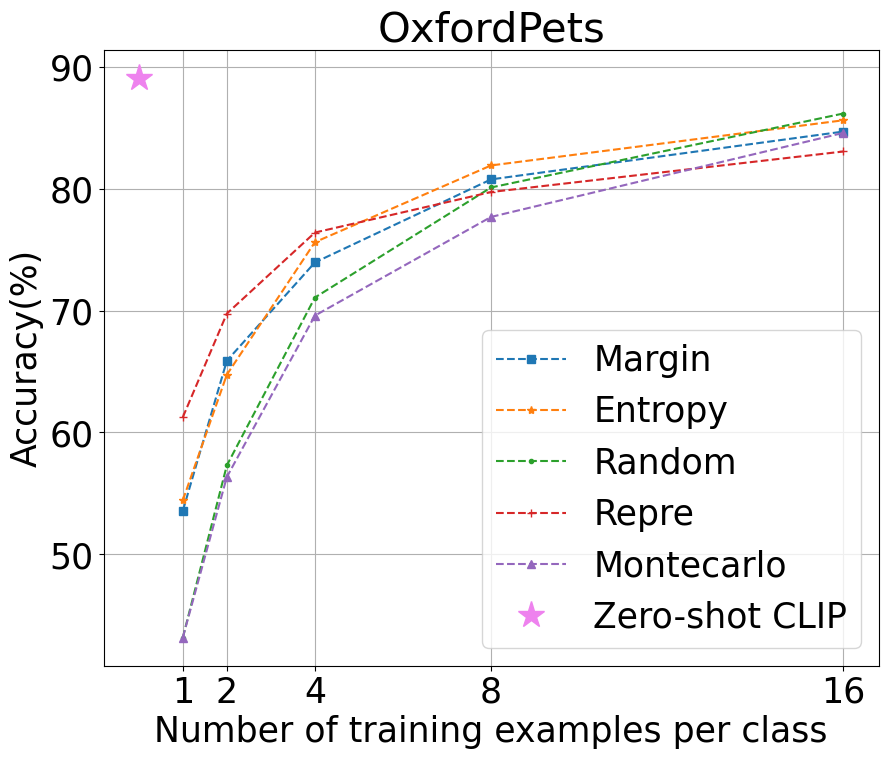} \\
\includegraphics[width=0.3\textwidth]{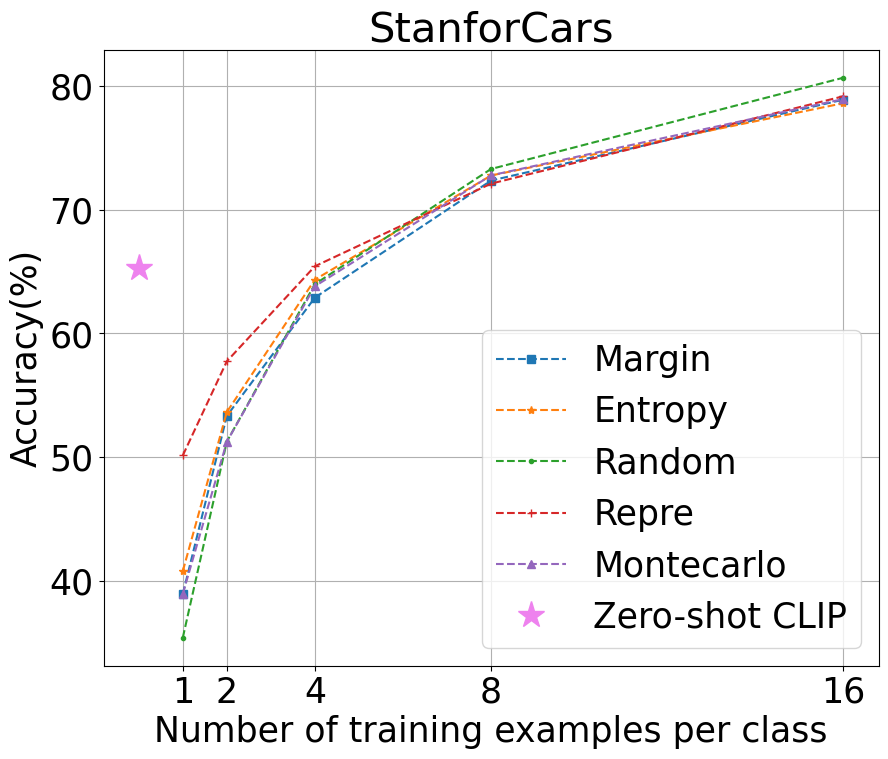} &
\includegraphics[width=0.3\textwidth]{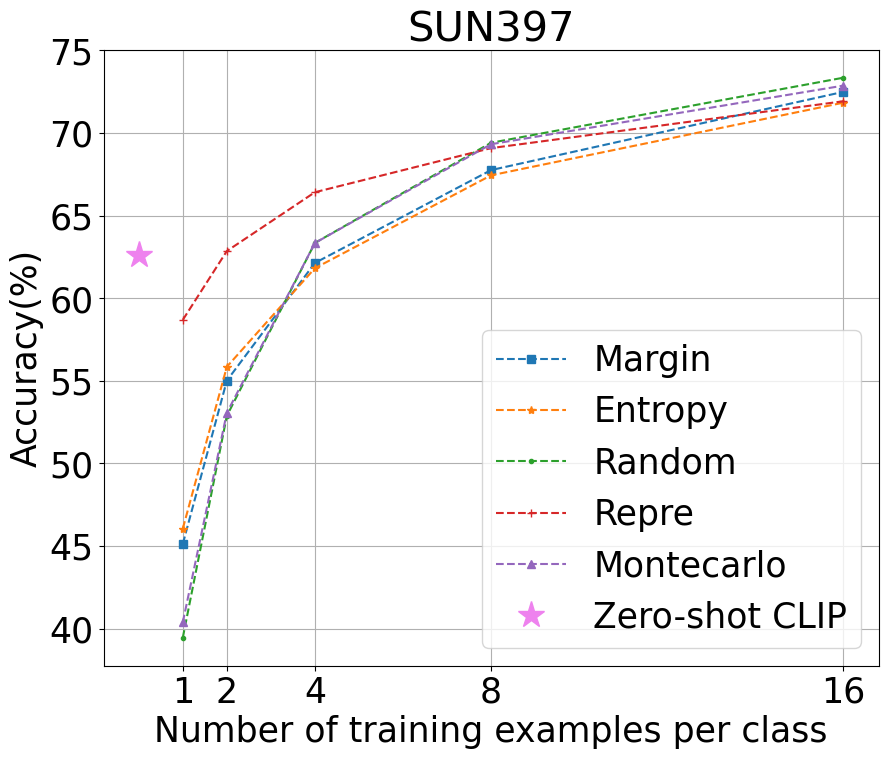} &
\includegraphics[width=0.3\textwidth]{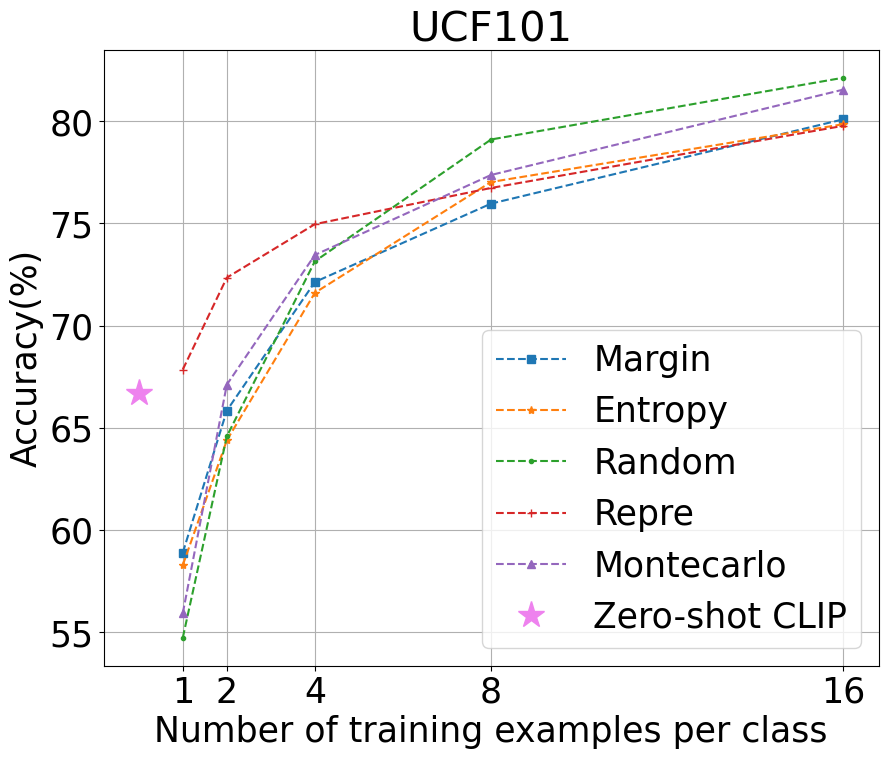} \\
\end{tabular}
\captionof{figure}{Main results of few-shot learning on the 11 datasets using the Linear probe approach. The results display the variability in performance across different datasets, illustrating the challenges and potential of the Linear probe in few-shot settings.}
\label{fig:linear}
\end{minipage}

\end{document}